\documentclass[preprint,12pt]{elsarticle}
\journal{Pattern Recognition}

\usepackage{xcolor}
\usepackage{mdframed}
\usepackage[margin=3.8cm]{geometry}
\usepackage[textwidth=1.8cm,textsize=scriptsize]{todonotes}
\usepackage{latexsym,amsmath,amsthm,amssymb,amsbsy,bbm,bm}
\usepackage{soul}
\usepackage{amssymb}
\usepackage{amsthm}
\usepackage{lineno}
\usepackage{soul}
\usepackage{color}
\newcommand{\red}[1]{\textcolor{red}{#1}}

\usepackage{siunitx} 
\usepackage{textcomp}
\usepackage{gensymb} 
\usepackage{booktabs}
\usepackage{multirow}
\newcolumntype{C}{>{\centering\arraybackslash}p{3.25em}}
\newcommand{\Real}{\mathbb R}
\newcommand{\bone}{{\mathbf 1}}
\newcommand{\bff}{{\mathbf f}}
\newcommand{\bfx}{{\mathbf x}}
\newcommand{\bfz}{{\mathbf z}}
\newcommand{\bfw}{{\mathbf w}}
\newcommand{\bfs}{{\mathbf s}}
\newcommand{\sfx}{{\mathsf x}}
\newcommand{\sfz}{{\mathsf z}}
\newcommand{\sfw}{{\mathsf w}}
\newcommand{\sfy}{{\mathsf y}}
\newcommand{\sfs}{{\mathsf s}}
\newcommand{\bfy}{{\mathbf y}}
\newcommand{\cN}{{\mathcal N}}
\newcommand{\cS}{{\mathcal S}}
\newcommand{\cR}{{\mathcal R}}
\DeclareMathOperator*{\argmin}{arg\,min}
\DeclareMathOperator*{\Tr}{Tr}
\DeclareMathOperator*{\Cov}{Cov}
\newcommand{\balpha}{\boldsymbol{\alpha}}
\newcommand{\bbeta}{\boldsymbol{\beta}}

\newcommand{\bSigma}{\boldsymbol{\Sigma}}
\newcommand{\Si}{{\boldsymbol{\Sigma}}}
\newcommand{\bPhi}{\boldsymbol{\Phi}}
\newcommand{\bPsi}{\boldsymbol{\Psi}}

\newcommand{\om}{{\bm{\omega}}}
\newcommand{\HH}{{\bf H}}
\newcommand{\II}{{\bf I}}
\newcommand{\K}{{\bf K}}
\newcommand{\R}{{\bf R}}
\newcommand{\LL}{{\bf L}}
\newcommand{\V}{{\bf V}}
\newcommand{\X}{{\bf X}}
\newcommand{\Y}{{\bf Y}}
\newcommand{\Z}{{\bf Z}}
\newcommand\independent{\protect\mathpalette{\protect\independenT}{\perp}}
\def\independenT#1#2{\mathrel{\rlap{$#1#2$}\mkern2mu{#1#2}}}
\newcommand{\GP}[0]{\mathcal{GP}} 
 
\newtheorem{proposition}{Proposition}
\theoremstyle{remark}
\newtheorem*{remark}{Remark}
\theoremstyle{definition}

\usepackage{hyperref}
\hypersetup{
    bookmarks=true,         % show bookmarks bar?
    unicode=false,          % non-Latin characters in Acrobat’s bookmarks
    pdftoolbar=true,        % show Acrobat’s toolbar?
    pdfmenubar=true,        % show Acrobat’s menu?
    pdffitwindow=false,     % window fit to page when opened
    pdfstartview={FitH},    % fits the width of the page to the window
    pdftitle={HSICGP},    % title
    pdfauthor={Camps et al},     % author
    pdfsubject={paper},   % subject of the document
    pdfcreator={Camps et al},   % creator of the document
    pdfproducer={Camps et al}, % producer of the document
    pdfkeywords={keyword1} {key2} {key3}, % list of keywords
    pdfnewwindow=true,      % links in new window
    colorlinks=true,       % false: boxed links; true: colored links
    linkcolor=blue,          % color of internal links (change box color with linkbordercolor)
    citecolor=blue,        % color of links to bibliography
    filecolor=blue,      % color of file links
    urlcolor=blue           % color of external links
}

\begin{document}

\begin{frontmatter}
\title{Kernel Dependence Regularizers and Gaussian Processes \\with Applications to Algorithmic Fairness}
\author{Zhu Li\fnref{addressOX}}
\author{Adrian Perez-Suay\fnref{addressUV}}
\author{Gustau Camps-Valls\fnref{addressUV}}
\author{Dino Sejdinovic\fnref{addressOX}}
\address[addressOX]{Department of Statistics, University of Oxford, 24-29 St Giles', Oxford OX1 3LB (UK)}
\address[addressUV]{Image Processing Laboratory (IPL), Universitat de Val\`encia, Catedr\'atico A. Escardino, 46980 Paterna, Val\`encia (Spain)}

\begin{abstract}
%\red{rewrite: Dino and Gustau}
Current adoption of machine learning in industrial, societal and economical activities has raised concerns about the fairness, equity and ethics of automated decisions. Predictive models are often developed using biased datasets and thus retain or even exacerbate biases in their decisions and recommendations. Removing the sensitive covariates, such as gender or race, is insufficient to remedy this issue since the biases may be retained due to other related covariates. We present a regularization approach to this problem that trades off predictive accuracy of the learned models (with respect to biased labels) for the fairness in terms of statistical parity, i.e. independence of the decisions from the sensitive covariates. In particular, we consider a general framework of regularized empirical risk minimization over reproducing kernel Hilbert spaces and impose an additional regularizer of dependence between predictors and sensitive covariates using kernel-based measures of dependence, namely the Hilbert-Schmidt Independence Criterion (HSIC) and its normalized version. This approach leads to a closed-form solution in the case of squared loss, i.e. ridge regression. Moreover, we show that the dependence regularizer has an interpretation as modifying the corresponding Gaussian process (GP) prior. As a consequence, a GP model with a prior that encourages fairness to sensitive variables can be derived, allowing principled hyperparameter selection and studying of the relative relevance of covariates under fairness constraints. Experimental results in synthetic examples and in real problems of income and crime prediction illustrate the potential of the approach to improve fairness of automated decisions.
\end{abstract}

\end{frontmatter}

%\modulolinenumbers[1]
%\linenumbers

\section{Introduction}

%\red{rewrite: Dino and Gustau}
%\begin{itemize}
%    \item Regularization is key in ML and in KMs in general
%    \item HSIC for GP reg
%    \item Two examples: fairness + coherent regression
%    \item Outline
%\end{itemize}

%Fairness is important - prior work overview.
%A na\"ive approach would be simply to leave sensitive variables out of analysis - this however is not a solution as other variables could be strongly correlated to the sensitive variables (give a running example), making predictions still unfair.
%Can project data to a subspace orthogonal to sensitive variables -- can be too restrictive as it ignores the problem we are trying to solve in the first place. Hence introduce the fairness penalty such that relevant criterion is jointly minimized with the dependence with sensitive variables.
%Discuss that the training set itself is unfair as there are potentially biases in the labelling -- a function which produced the labels is not necessarily the function we wish to learn - so we are solving a problem beyond the classical empirical risk minimization. What are the implications of this?

%% GCV:
\subsection{Motivation}
Current and upcoming pervasive application of machine learning algorithms promises to have an enormous impact on people's lives. %, and impact decisions on education, economy, health care, and climate policies. 
%New social and economic activities massively exploit big data and machine learning algorithms to do inferences, 
%Curricula screening, wage determination, and loan risk assessment are just a few examples. Governments and institutions passed laws and regulations to endorse machines with fairness, equity and ethics to treat these problems. 
For example, algorithms now decide on the best curriculum to fill in a position~\citep{hoffman2017discretion}, determine wages~\citep{dieterich2016compas}, help in pre-trial risk assessment~\citep{brennan2009evaluating}, 
and evaluate risk of violence~\citep{cunningham2006actuarial}. Concerns were raised about the lack of fairness, equity and ethics in machine learning to treat these types of problems\footnote{See for example the \href{http://fra.europa.eu/en/publication/2011/handbook-european-non-discrimination-law}{Handbook on European non-discrimination law} and the Paycheck Fairness Act in the \href{http://www.usdoj.gov}{U.S. Federal Legislation}.}.
%\footnote{Workshop on \href{http://www.fatml.org/}{Fairness, Accountability, and Transparency in Machine Learning}}. 
Indeed, standard machine learning models are far from being fair, just, or equitable: they will retain and often exacerbate systemic biases present in data. For example, a model trained simply to minimize a loss with respect to human-provided labels which are subject to a cognitive bias cannot be expected to be free from that bias. More nuanced modelling approaches are needed to move towards fair decision-making processes based on machine learning algorithms. %is undoubtedly a timely and important concern.
New algorithms should also be easy to use, implement and interpret.

\subsection{Approaches to Fairness in Machine Learning}

Fairness is an elusive concept, and adopts many forms and definitions. The field is vast, and a wide body of literature and approaches exists \citep{Pedreschi:2008, kamiran2009classifying,chouldechova2018frontiers}.
%We frame our specific contribution in the particular field of indirect discrimination for treating disparate impact problems. In this context, outcomes should not differ based on individuals' protected (sensitive) information or group membership. This problem has been addressed by defining {\em ad hoc} classification rules~\citep{Pedreschi:2008,Ruggieri:2010} or preprocessing the data to remove sensitive dependencies explicitly~\citep{kamiran2009classifying,Luo2015, Feldman:2015,Ristanoski2013}. Down-weighting sensitive features or directly removing them have been the preferred choices~\citep{zeng2015interpretable}. 
%Simply removing the sensitive covariates (such as gender, disability, or race, to predict, e.g., monthly income or credit score) is, however, often insufficient as related variables (from which a sensitive covariate may be inferred) may enter the model anyway and the bias is retained. Including covariates related to the sensitive variables in the models is called {\em redlining}, and induces the well-known problem known as the {\em omitted variable bias} (OVB)~\citep{clarke2005phantom}. Other authors have focused on finding transformations of the input space in order to extract features that do not retain information about the sensitive input variables~\citep{ZemelICML13}.
%There are two main notions: individual fairness and group fairness. 

Let us broadly distinguish into two classes of fairness: individual and group fairness. 
On one hand, \emph{individual fairness} \citep{dwork2012fairness,joseph2016fairness,kim2018fairness,heidari2018fairness} is a notion that can be roughly understood as: ``similar individuals should be treated similarly'. An example is \cite{dwork2012fairness}, where it is assumed that there exists a similarity measure among individuals, and the goal is to find a classifier that returns similar outcomes for individuals with high similarity. \citet{joseph2016fairness} gives another formalization of individual fairness, which can be loosely described as: ``less qualified individual should not be favoured'', where the notion of quality is estimated from data. Although practically important, there are certain obstacles that prevent individual fairness being widely adopted in practice. For example, the approach from \cite{dwork2012fairness} requires a pre-agreed similarity measure which may be difficult to define. Also, employing individual fairness requires evaluation on any pair of individuals in the dataset and when dealing with large datasets, such computation may be infeasible.

On the other hand, \emph{group fairness} focuses on the inequality at the group level (where groups may be defined using a sensitive variable such as race or gender). More broadly, outcomes should not differ systematically based on individuals' protected (sensitive) information or group membership. This problem has been addressed by modifying classification rules~\citep{Pedreschi:2008,Ruggieri:2010} or preprocessing the data to remove sensitive dependencies explicitly~\citep{kamiran2009classifying,luo2015discrimination, feldman2015certifying,Ristanoski2013}. Down-weighting sensitive features or directly removing them have been proposed \citep{zeng2015interpretable}. However, simply removing the sensitive covariates (such as gender, disability, or race, to predict, e.g., monthly income or credit score) is often insufficient as related variables may still enter the model. Sensitive covariate may be inferred from those related variables and the bias is retained. Including covariates related to the sensitive variables in the models is called {\em redlining}, and induces the problem known as the {\em omitted variable bias} (OVB). Alternative approaches seek \emph{fair representation learning}, i.e. achieving fairness through finding an optimal way to preprocess the data and map it into a latent space where all information about the sensitive variables is removed. After such preprocessing, standard machine techniques are employed to build predictive models. Examples of these methods include \cite{ZemelICML13,kamiran2012data,adebayo2016iterative,calmon2017optimized}.  %kamiran2009classifying,kamiran2010classification
Statistical parity approaches, on the other hand, directly impose the independence between predictor and sensitive variables \citep{calders2010three,kamishima2012fairness,feldman2015certifying,perez2017fair}. Various other statistical measures across groups can be considered, e.g. \emph{equalized odds} which require that the false positive and false negative rates should be approximately equal across different groups \citep{kleinberg2016inherent,hardt2016equality,Chouldechova17,zafar2017fairness}, and other examples are given in  \cite{berk2018fairness}. Group fairness is attractive because it is simple to implement, it often leads to convex optimization problems, and it is easy to verify in practice. However, as argued by \cite{dwork2012fairness}, group fairness cannot give guarantees to individuals as only average fairness to members of each subgroup is attained.

\subsection{Regularization for Group Fairness}
In this paper, we build on the work of \cite{perez2017fair} which falls within the framework of group fairness and was the first work that considered the notion of statistical parity with continuous labels. In particular, independence between predictor and sensitive variables is imposed by employing a kernel dependence measure, namely the Hilbert-Schmidt Independence Criterion (HSIC) \cite{Gretton05}, as a \emph{regularizer} in the objective function. Regularization is one of the key concepts in modern supervised learning, which allows imposing structural assumptions and inductive biases onto the problem at hand. It ranges from classical notions of sparsity, shrinkage, and model complexity to the more intricate regularization terms which allow building specific assumptions about the predictors into the objective functions, e.g. smoothness on manifolds \citep{Belkin2006}. Such regularization viewpoint for algorithmic fairness was presented in~\cite{kamishima2012fairness} in the context of classification, and was extended to regression and dimensionality reduction with kernel methods in~\cite{perez2017fair}. Our work extends \cite{perez2017fair} in the following three ways. Firstly, we give a general framework of empirical risk minimization with \emph{fairness regularizers} and their interpretation. Secondly, we derive a Gaussian Process (GP) formulation of the fairness regularization framework, which allows uncertainty quantification and principled hyperparameter selection. Finally, we introduce a normalized version of the fairness regularizer which makes it less sensitive to the choice of kernel parameters.
We demonstrate how the developed fairness regularization framework trades off model's predictive accuracy (with respect to potentially biased data) for independence to the sensitive covariates. It is worth noting that, in our setting, a function which produced the labels is not necessarily the function we wish to learn, so that the predictive accuracy is not necessarily a gold-standard criterion.

The paper is structured as follows. The general framework, together with the relevant background, is developed in \S2. In \S3, we develop a Gaussian Process (GP) interpretation of kernel dependence regularization. We give some instances of fairness-regularized ERM and their interpretation in \S4. \S5 describes the normalized kernel dependence regularizer. Experimental results are presented in \S6. % in toy examples and prediction of income and credit risk assignment [!and crime!] under gender and race discrimination illustrate the usefulness of the approach. 
We conclude the work with some remarks and further work in \S7.

%%%%%%%%%%%%%%%
%%%%%%%%%%%%%%%
%%%%%%%%%%%%%%%
%%%%%%%%%%%%%%%
%%%%%%%%%%%%%%%

\section{Regression with dependence penalization}

\iffalse

\red{rewrite: Michael, Dino. to avoid focus on fairness, which should be just an example here. If we go for NatComm instead, all the maths section must go at the end. Discuss.}

\begin{itemize}
    \item the choice of a dependence measure, define HSIC, discuss normalized version and other choices
    \item importance of using a nonlinear kernel on sensitive variables
    \item importance of using a nonlinear kernel on predicted responses -- example of income vs gender where expectation is the same but variance is larger for men. discuss if this is a valid notion of fairness, i.e. top 10\% earners are all men (but also bottom 10\%)
\end{itemize}
\fi

%%%%% The reg framework

\subsection{Fairness regularization framework}
We build on a pragmatic definition of fairness following~\citep{Chouldechova17}. We are given a set of inputs, $\bfx_i\in{\mathcal X}$, and the corresponding targets, $y_i\in{\mathcal Y}$, for $i=1,\ldots,n$. Furthermore, we have observations of sensitive inputs $\bfs_i\in{\mathcal S}$ (sensitive inputs $\bfs_i$ could be treated as a subset of $\bfx_i$). We take $\bfx_i$ to be an iid sample from an $\mathcal X$-valued random variable $\sfx$, and similarly for $\sfs$. For simplicity, we will assume that the inputs are vectorial, i.e. $\mathcal X\subseteq\Real^{d\times 1}$, $\mathcal S\subseteq\Real^{q\times 1}$ and that the targets are scalar, i.e. ${\mathcal Y}\subseteq \Real$, but the exposition can be trivially extended to non-Euclidean or structured domains which admit positive definite kernel functions. We let $\X\in\Real^{n\times d}$ denote the matrix of $n$ observed inputs corresponding to $d$ explanatory covariates, $\mathbf{S}\in\Real^{n\times q}$ denotes the set of $q$ sensitive (protected) variables, $\bfy\in\Real^{n\times 1}$ denotes the vector of observed targets, which we assume are corrupted with historical biases, and $\hat \bfy$ is the predictor. We will also introduce the following notions of fair predictors in terms of statistical parity. The fitted predictor $f:{\mathcal X}\to{\mathcal Y}$ is said to be \emph{parity-fair} to the sensitive input $\sfs$ if and only if $f(\sfx)$ %predictions $\hat{\y}$ are 
is statistically independent of $\sfs$. Moreover, it is said to be \emph{parity-fair in expectation} to the sensitive input $\sfs$ if and only if $\mathbb E_{\sfx|\sfs}\left[f(\sfx)|\sfs\right]$ does not depend on $\sfs$.

\begin{remark}
We note that parity-fairness implies parity-fairness in expectation but that the converse is not true. For example, it may be possible that the conditional variance $\text{Var}[f(\sfx)|\sfs]$ still depends on $\sfs$. For a concrete example, consider the case of modelling income where gender is a sensitive variable. Parity-fairness in expectation implies that the mean predicted income does not depend on the gender, but it is possible that, e.g. the variance is larger for one of the genders. This is hence a weaker notion of fairness, as it may still result in predictions where, say, the top 10\% earners all have the same gender. 
\end{remark}
%since we are always using a linear kernel on $f(\sfx)$, we need to introduce an appropriate notion of fairness!
Fitting a fairness-regularized predictor $f_\ast\in{\mathcal H}$ for some hypothesis class $\mathcal H$, reduces to optimizing a regularized empirical risk functional ~\citep{kamishima2012fairness,perez2017fair}:
\begin{equation}\label{eq:functional}
f_\ast = \argmin_{f\in\mathcal H} \frac{1}{n} \sum_{i=1}^n V(f(\bfx_i),y_i) + \Omega(f) + \eta I(f(\sfx),\sfs) 
\end{equation}
%\begin{equation}\label{eq:functional}
%{\mathcal L} = V(f,\bfy) + \lambda~\Omega(f) + \eta~I(f,\mathbf{S}),
%\end{equation}
where $V$ is the loss function, $\Omega$ acts as an overfitting/complexity penalty on $f$, and $I$ measures the statistical dependence between the model $f$ and the protected variables. By setting $\eta=0$, standard, yet potentially biased, machine learning models are obtained. 

%ds: the below seemed to conflate the loss (logistic regression) with the model (RKHS function), so I've casted (1) as ERM and changed slightly
The framework admits many variants depending on the loss function $V$, regularizer $\Omega$ and the dependence measure, $I$. In~\citep{kamishima2012fairness}, a logistic loss was used and $I$ was a simplified version of the mutual information estimator. In~\citep{perez2017fair}, the hypothesis class was a reproducing kernel Hilbert space (RKHS), and the dependence measure $I$ was Hilbert-Schmidt Independence Criterion (HSIC), based on the norm of the particular cross-covariance operator on RKHSs~\citep{Gretton05}, allowing one to deal with several sensitive variables simultaneously. When combined with the framework of kernel ridge regression, a closed-form solution is obtained. In this paper, we extend the latter formalism and introduce a Gaussian process (GP) treatment of the problem. Then we study the HSIC penalization as a modified GP prior, and explore the aspects of HSIC normalization, and the interpretability of the hyperparameters inferred under the GP framework. Before that, let us fix notation and review the basics of GP modeling and kernel-based dependence measures. 

\subsection{GP models}
%We are given a set of input points $\bfx_i\in\Real^d$, with corresponding labels $y_i\in\Real$, $i=1,\ldots,n$. 
In GP modeling, observations $y_i$ are assumed to arise from a probabilistic model $p_\lambda\left(y_i|f(\bfx_i)\right)$, parametrized by the evaluation $f(\bfx_i)$ of a latent function $f$ at the input $\bfx_i$. Here, $\lambda>0$ is an optional hyperparameter used to rescale the log-likelihood, i.e. $\log p_\lambda\left(y_i|f(\bfx_i)\right) = \text{const}+\frac{1}{\lambda}\log p\left(y_i|f(\bfx_i)\right)$.  For example, in GP regression, we assume a normal likelihood, i.e. $\log p_\lambda(y_i|f(\bfx_i))=\text{const}-\frac{1}{2\lambda}(y_i-f(\bfx_i))^2$. Equivalently, the latent function is impaired by a Gaussian noise of variance $\lambda$, i.e. $y_i = f(\bfx_i) + \varepsilon_i$, $\varepsilon_i\sim{\mathcal N}(0,\lambda)$, independently over $i=1,\ldots,n$.   %, and the inference proceeds in a Bayesian, non-parametric way.
A Gaussian process prior, typically zero-mean\footnote{For example, in regression, it is customary to subtract sample average from the targets $\{y_i\}_{i=1}^n$, and then to assume a zero-mean model.},  is placed on the latent function $f$, denoted $f(\bfx)\sim \mathcal{GP}(0, k_{\boldsymbol{\theta}}(\bfx,\bfx'))$, where $k_{\boldsymbol{\theta}}(\bfx,\bfx')$ is a covariance function parametrized by $\boldsymbol{\theta}$. % and $\sigma^2$ is a hyperparameter that specifies the noise power.
%Instead of proposing a parametric form for $f$ and learning its parameters in order to fit observed data well, 
Advantageously, GPs provide a coherent framework to select model hyperparameters $\boldsymbol{\theta}$ and $\lambda$ by maximizing the marginal log-likelihood, or to pursue Bayesian treatment of hyperparameters. Moreover, they yield a posterior distribution over predictions $f(\bfx_\star)$ for new inputs $\bfx_\star$, allowing to quantify uncertainty and return a predictive posterior of target $y_\star$, not just a point estimate.
We will denote latent function evaluations over all inputs as $\bff=\left[f(\bfx_1),\ldots,f(\bfx_n)\right]^\top$. 

\subsection{Dependence measures with kernels}

%DS: TODO, describe and define cross-covariance operator!

Consider random variables $\sfz$ and $\sfw$ taking values in general domains $\mathcal Z$ and $\mathcal W$. Given kernel functions $m$ and $l$ on $\mathcal Z$ and $\mathcal W$ respectively, with RKHSs $\mathcal H_m$ and $\mathcal H_l$, the cross-covariance operator is defined as a linear operator $\Sigma_{\sfz\sfw}:\mathcal H_l \to \mathcal H_m$ such that $\langle g,\Sigma_{\sfz\sfw}h\rangle_{\mathcal H_m}=\Cov[g(\sfz),h(\sfw)]$, for all $g\in\mathcal H_m$, $h\in\mathcal H_l$. Hilbert-Schmidt Independence Criterion (HSIC) measuring dependence between $\sfz$ and $\sfw$ is then given by the Hilbert-Schmidt norm of $\Sigma_{\sfz\sfw}$.
HSIC can be understood as a maximum mean discrepancy (MMD) \citep{Gretton2012} between the joint probability measure of $\sfz$ and $\sfw$ and the product of their marginals. Given the dataset ${\mathcal{D}}$ with $n$ pairs drawn from the joint $P(\bfz,\bfw)$, an empirical estimator of HSIC is defined as~\citep{Gretton05}:
\begin{eqnarray}
\widehat{\text{HSIC}}_{m,l}(\sfz,\sfw)= %\dfrac{1}{n^2} \text{Tr}({\bf K}_x{\bf H}{\bf K}_y{\bf H}) = 
\dfrac{1}{n^2} \Tr({\bf M}{\bf H}{\bf L}{\bf H}),
\label{empHSIC}
\end{eqnarray}
%where $\text{Tr}(\cdot)$ is the trace operation (the sum of the diagonal entries), 
where ${\bf M}$, ${\bf L}$ are the kernel matrices computed on observations $\{\bfz_i\}_{i=1}^n$ and $\{\bfw_i\}_{i=1}^n$ using kernels $m$ and $l$ respectively, and ${\bf H}= {\bf I} - \frac{1}{n}\mathbbm{1}\mathbbm{1}^\top$ has the role of centering the data in the feature space. %s ${\mathcal F}$ and ${\mathcal G}$, respectively. 
%Here $\delta$ represents the Kronecker symbol, where $\delta_{i,j}=1$ if $i=j$, and zero otherwise.
For a broad family of kernels $m$ and $l$ (including e.g. Gaussian RBF and Mat\'ern family), the population HSIC equals 0 if and only if $\sfz$ and $\sfw$ are statistically independent, cf. \citep{Gretton05}. Hence, nonparametric independence tests consistent against all departures from independence can be devised using HSIC estimators with such kernels. 
Note, however, that the selection of the kernel functions and their parameters have a strong impact on the \emph{value} of HSIC estimator. As we will see, this is important when HSIC is used as a regularizer, as it generally leads to different predictive models.

Moreover, HSIC is sensitive to the scale appearing in the marginal distributions of $\sfz$ and $\sfw$ and their units of measurements and hence needs an appropriate normalization if it is to depict a dependence measure useful for, e.g. relative dependence comparisons. This problem is well recognized in the literature and a normalized version of HSIC, called NOCCO (NOrmalized Cross-Covariance Operator) was introduced in \citep{Fukumizu:NIPS2007}.

\section{Interpretations of HSIC penalization}
%\red{rewrite: Michael, Dino and Gustau. Should be a differnet section or a subsec?}

%\paragraph{Bayesian formulation}
Consider a particular instantiation of the regularized functional in \eqref{eq:functional} given by %with HSIC between the function and the sensitive variables being the fairness penalty term
\begin{equation}
    \min_{f\in\mathcal H_k} \bigg\{\frac{1}{n}\sum_{i=1}^n V(f(\bfx_i),y_i) + \frac{\lambda}{n} \Vert f \Vert _{\mathcal H_k}^2 +
    \eta \widehat{\text{HSIC}}_{m,l}(f(\sfx),\sfs)\bigg\},\label{fairlearning_hsic}
\end{equation}
where we adopted the reproducing kernel Hilbert space (RKHS) $\mathcal H_k$ as a hypothesis class and added a fairness penalization term consisting of an estimator of HSIC between the predicted response $f(\sfx)$ and the sensitive variable $\sfs$.

With appropriate choices of kernels $m$ and $l$, HSIC regularizer captures all types of statistical dependence between $f(\sfx)$ and $\sfs$. However, we will here focus on \emph{fairness in expectation} as it will give us a convenient link to GP modelling. Fairness in expectation corresponds to adopting a linear kernel on $f(\sfx)$, i.e., $m(f(\bfx_i),f(\bfx_j))=f(\bfx_i)f(\bfx_j)$. Estimator \eqref{empHSIC} then simplifies to 
\begin{equation}
    \widehat{\text{HSIC}}_{m,l}(f(\sfx),\sfs)=\frac{1}{n^2}\Tr(\bff\bff^\top\HH\LL \HH)=\frac{1}{n^2}\bff^\top \HH \LL \HH\bff.
\end{equation} Given that this fairness penalty term only depends on the unknown function$f$ through its evaluations $\bff$ at the training inputs $\{\bfx_i\}$, direct application of Representer theorem \citep{kimeldorf1970} tells us that the optimal solution can be written as $f=\sum_{i=1}^n\alpha_i k(\cdot,\bfx_i)$. 
Hence, we obtain the so called \emph{dual} problem
\begin{equation}\label{regdual}
    \min_{\balpha \in \Real^n} \bigg\{ \frac{1}{n}\sum_{i=1}^n V(f(\bfx_i),y_i) + \frac\lambda n \balpha^\top \K \balpha + \frac \eta {n^2} \balpha^\top \K\HH\LL\HH\K \balpha\bigg\}.
\end{equation}
The problem \eqref{regdual} can now be solved for $\balpha$ directly, and in the case of squared loss, it has a closed form solution \citep{perez2017fair}.

%Here, $\HH={\bf I}-\frac{1}{n}\bone\bone^\top$ is the centering matrix

\subsection{Modified Gaussian Process Prior}
For a Bayesian interpretation of \eqref{fairlearning_hsic}, we here assume that the loss corresponds to the negative conditional log-likelihood in some probabilistic model, i.e. that
$V(f(\bfx_i),y_i)=-\log p\left(y_i|f(\bfx_i)\right)$, which is true for a wide class of loss functions. Hence, we will write \eqref{fairlearning_hsic} as: 
\begin{equation}
    \min_{f\in\mathcal H_k} \bigg\{-\sum_{i=1}^n \frac{\log p(y_i|f(\bfx_i))}{\lambda} + \Vert f \Vert _{\mathcal H_k}^2 +
    \delta \bff^\top \HH \LL \HH\bff \bigg\},\label{fairbayesianlearning_hsic}
\end{equation}
%where $\lambda$ is treated as the scale hyperparameter of the likelihood, e.g. as the variance of noise in GP regression where $\log p(y_i|f(\bfx_i))=-\frac{1}{2}(y_i-f(\bfx_i))^2$, and 
where we write $\delta=\eta/\lambda n$ (note that the objective \eqref{fairlearning_hsic} is rescaled by $n/\lambda$ such that the regularization parameter $\lambda$ now plays the role of rescaling the log-likelihood).

Consider now using explicit feature mapping $\bfx_i\mapsto \phi(\bfx_i)$ (for the moment assumed finite-dimensional) and denoting the feature matrix by $\bPhi$, we have $\bff=\bPhi\bbeta$ and thus can recast optimization as (so called \emph{primal} problem) with some abuse of notation\footnote{We write $V(\bfy,\bff)=-\sum_{i=1}^n \log p(y_i|f(\bfx_i))$ to denote the rescaled conditional negative log-likelihood.}:
\begin{equation}\label{regprimal}
    \min_{\beta \in \Real^m} \bigg\{\frac{1}{\lambda}V(\bfy,\bPhi\bbeta) 
    + \bbeta^\top \bbeta 
    + \delta \bbeta^\top \bPhi^\top \HH\LL\HH \bPhi \bbeta\bigg\}.
\end{equation}
These problems give us an insight about how the two regularization terms interact. It is well known that solutions to regularized ERM over RKHS $\mathcal H_k$ are closely related to GP models using covariance kernel $k$ -- for a recent overview, cf. \citep{kanagawa2018gaussian} and references therein. In particular, by inspecting \eqref{regprimal}, the two regularization terms correspond, up to an additive constant, to a negative log-prior of $\bbeta\sim\cN\left(0,\left( \II+\delta \bPhi^\top \HH\LL\HH\bPhi\right)^{-1}\right)$, which in turn gives a prior on the evaluations $\bff\sim\cN\left(0,\bPhi\left(\II + \delta \bPhi^\top \HH\LL\HH\bPhi\right)^{-1}\bPhi^\top\right)$. By directly applying the Woodbury-Morrison formula, the covariance matrix in this prior becomes $(\K^{-1}+\delta \HH\LL\HH)^{-1}$, compared to $\K$ in the standard GP case. Thus, adding an HSIC regularizer corresponds to modifying the prior on function evaluations $\bff$. A natural question arises:

\begin{mdframed}[backgroundcolor=gray!20] \emph{Question 1}: can the fairness-regularized ERM in \eqref{fairlearning_hsic} be interpreted as simply modifying the GP prior on the whole function $f$ into a \emph{fair GP prior}?
\end{mdframed}

As the next proposition shows, the answer to \emph{Question 1} is positive. The proof is given in \ref{gp_conversion_proof}. 

\iffalse
***$\lambda$ is generally the scale hyperparameter of the likelihood, not of the prior -- so this needs explanation, e.g. in the context of ridge regression***
***Potential issue is that $HLH$ is not an invertible matrix, since $HLH\bone=0$.***
\fi

\begin{proposition}\label{gp_conversion}
Solution to \eqref{fairbayesianlearning_hsic} corresponds to the posterior mode in a Bayesian model using a modified GP prior 
\begin{equation}
f \sim \GP\left(0, k(\cdot,\cdot)-k_{\X\cdot}^\top({\bf K}\HH\LL\HH+ \delta^{-1}\II)^{-1}\HH\LL\HH k_{\X\cdot}\right).    
\end{equation}
where $k_{\X\cdot} = [k(\cdot,\bfx_1),\cdots,k(\cdot,\bfx_n)]^\top$, for any training set $\{\bfx_i\}_{i=1}^n$.
\end{proposition}

Several important consequences of the GP interpretation will allow us to improve the fair learning process. In particular, the GP treatment allows us to easily derive uncertainty estimates and perform hyperparameter learning using marginal log-likelihood maximization, which is more practical than typical cross-validation strategy limited to simple parameterizations. More importantly, appropriate inference of the model (parameters and hyperparameters) thus yield closer insight into the fairness tradeoffs.

\subsection{Projections using Cross-Covariance Operators}
We can  derive an additional intepretation of the fairness regularizer in terms of cross-covariance operators. Namely, by considering an explicit feature map $\bfs_i\mapsto \psi(\bfs_i)$ corresponding to the kernel $l$, and denoting the feature matrix by $\bPsi$, i.e. $\LL=\bPsi\bPsi^\top$ we see that the fairness regularizer in \eqref{regdual} reads
\begin{equation}
    \bbeta^\top \bPhi^\top \HH\LL\HH \bPhi \bbeta = \Vert \bPsi^\top \HH \bPhi \bbeta\Vert_2^2 = n^2 \Vert \hat\bSigma_{\sfs\sfx} \bbeta\Vert_2^2,
\end{equation}
where $\hat\bSigma_{\sfs\sfx}=\frac{1}{n}\bPsi^\top\HH\bPhi $ is the empirical cross-covariance matrix between feature vectors $\phi(\bfx_i)$ and $\psi(\bfs_i)$.
This interpretation also holds in the case of infinite-dimensional RKHSs $\mathcal H_k$ and $\mathcal H_l$. For an infinite-dimensional version of primal formulation, we define sampling operator $\cS:\mathcal H_k \to \Real^n$, $\cS f=\bff$. Then the HSIC regularizer becomes
$$\bff^\top \HH \LL \HH\bff =\langle \cS f, \HH\LL\HH\cS f\rangle_{\Real^n} = \langle f, \cS^* \HH\LL\HH\cS f\rangle_{\mathcal H_k},$$
where the adjoint $\cS^*$ acts as $\cS^*:\alpha\mapsto \sum_{i=1}^n\alpha_i k(\cdot,x_i)$. Moreover, if we define similarly the sampling operator for kernel $l$, i.e. $\cR:\mathcal H_l\to\Real^n$, $\forall h \in \mathcal{H}_l$ with $\cR h = [h(s_1),\ldots,h(s_n)]^\top$, then $\LL=\cR\cR^*$ and $\cS^*\HH\cR=n\hat\bSigma_{\sfx\sfs}$, $\cR^*\HH\cS=n\hat\bSigma_{\sfs\sfx}$. Here, $\hat\bSigma_{\sfx\sfs}: \mathcal H_l \to \mathcal H_k$ and $\hat\bSigma_{\sfx\sfs}: \mathcal H_k \to \mathcal H_l$  are the empirical cross-covariance operators \citep{Fukumizu:NIPS2007}, i.e. $\forall f\in\mathcal H_k, h\in\mathcal H_l$ 
$$\langle f,\hat\bSigma_{\sfx\sfs}h\rangle_{\mathcal H_k}=\langle \hat\bSigma_{\sfs\sfx}f,h\rangle_{\mathcal H_l}=\frac 1 n \sum _{i=1}^n f(x_i)h(s_i)-\frac 1 n \sum _{i=1}^n f(x_i)\frac 1 n \sum _{i=1}^n h(s_i).$$

Thus, the overall objective can be written as

\begin{equation}
    \min_{f\in\mathcal H_k} \bigg\{\frac{1}{\lambda}V(\bfy,\cS f) 
    + \left\langle f, \left(\II+ \delta n^2\hat\bSigma_{\sfx\sfs}\hat\bSigma_{\sfs\sfx}\right) f\right\rangle_{\mathcal H_k}\bigg\}.
\end{equation}

%so one can consider whether $\lambda \II + \eta \cS^* \HH\LL\HH\cS$ is an invertible operator on RKHS so that it defines a Gaussian measure?
Here, $\II$ denotes the identity on $\mathcal H_k$. Hence, the additional regularization term is up to scaling simply 
$$\langle f, \hat\bSigma_{\sfx\sfs}\hat\bSigma_{\sfs\sfx} f\rangle_{\mathcal H_k} = \Vert \hat\bSigma_{\sfs\sfx} f\Vert_{\mathcal H_l}^2 = \sum_{i\in I}\widehat{\text{Cov}}(u_i(\sfs),f(\sfx)),$$
where $u_i$ is an arbitrary basis of $\mathcal H_l$. This gives another insight into the fairness regularizer as an action of the empirical cross-covariance operator between sensitive and remaining inputs $\sfs$ and $\sfx$ on the learned function\footnote{Note that this operator is different from the cross-covariance operator defining HSIC in \eqref{fairlearning_hsic} itself, as the latter pertains to cross-covariance between $\sfs$ and $f(\sfx)$}. As we shall see, this perspective will also allow us to construct a normalized version of fairness regularizer in Section \S5.

%DS: TODO: maybe move below somewhere else.
%As we shall see, this GP interpretation gives us that the modified prior covariance corresponds to a posterior covariance for meta-observations encouraging that the function remains constant as the sensitive variables vary. 
 %Demonstrate an example with several hyperparameters so that cross-validation is impractical - e.g. using ARD kernel.

\section{Instances of dependence-regularized learning}

%\red{rewrite: Michael, Dino and Gustau. Should be a differnet section or a subsec? Make the comment of -HSIC and talk about consistency...}

%Recall that the Hilbert-Schmidt norm of $\hat{\bSigma}_{\sfx\sfs}$ is the HSIC estimator which quantifies the dependence between $\sfx$ and $\sfs$ .
In this section, we give two concrete examples of fair learning and give illustrations how the fairness penalty enforces the fairness in both the ridge regression setting and in the Bayesian learning setting. As before, we denote by $\bSigma_{\sfx\sfs}$ the cross-covariance operator and by $\hat{\bSigma}_{\sfx\sfs}$ its empirical version. 
%[!!!DS: TODO: here cross-covariance operator must depend linearly on $\sfx$ but need not depend linearly on $\sfs$ -- need to clarify that!!!]
\paragraph{Fair Linear Regression} We start with the simple case of linear regression. We note that the kernel $k$ on $\sfx$ is then linear, while the kernel $l$ on $\sfs$ need not be. For simplicity, let us assume that $l$ is finite-dimensional and write its explicit feature map as $\psi(\bfs)\in\mathbb R^m$. Thus, we have the following minimization problem:
\begin{eqnarray}%{rCl}
\bbeta_* ~ :&=& \argmin_{\bbeta} \frac{1}{n}\|\bfy-\X\bbeta\|_2^2 + \frac \lambda n \|\bbeta\|_2^2 +  \eta \|\hat{\bSigma}_{\sfs\sfx}\bbeta\|_2^2\nonumber \\
&=&\argmin_{\bbeta} \frac{1}{\lambda}\|\bfy-\X\bbeta\|_2^2 + \bbeta^\top(\II + \delta n^2\hat{\bSigma}_{\sfx\sfs}\hat{\bSigma}_{\sfs\sfx})\bbeta. \label{fair_regression} 
\end{eqnarray}
The purpose of fair linear regression is to predict $\bfy$ from inputs $\X$ while ensuring that the predictions are independent of the sensitive variable ${\sfs}$. From \eqref{fair_regression}, we see that the HSIC regularizer penalizes \emph{the weighted norm} of $\bbeta$. The weight on each dimension of $\bbeta$ is guided by the cross-covariance operator $\hat{\bSigma}_{\sfs\sfx}:\mathbb R^d \to \mathbb R^m$. As a result, if a dimension ${\sfx}_i$ in $\sfx$ has a high covariance with any of the entries in $\psi(\sfs)$, its corresponding coefficient $\beta_i$ will be shrank towards zero, leading to a low covariance between $f(\sfx)$ and $\psi(\sfs)$. This can be illustrated by the following toy case. Since feature spaces are finite-dimensional, we can treat $\hat{\bSigma}_{\sfs\sfx}$ as an $m\times d$ matrix. Say that $m < d$ and that $\hat{\bSigma}_{\sfs\sfx}$ has zero-off diagonal entries (i.e. the only non-zero cross-correlations are between the $i$-th dimension of $\sfx$ and the $i$-th dimension of $\psi(\sfs)$). We further enlarge $\hat{\bSigma}_{\sfs\sfx}$ to be a $d \times d$ matrix by appending zeros. We denote the diagonal elements of the enlarged matrix as $\sigma_1,\ldots,\sigma_d, \sigma_i \geq 0$.  As a result, $\hat{\bSigma}_{\sfx\sfs}\hat{\bSigma}_{\sfs\sfx} \in \mathbb{R}^{d\times d}$ is symmetric and diagonal with diagonal elements $\sigma_1^2,\ldots,\sigma_d^2$. Now the second term in Eq. \eqref{fair_regression} is simply:
\begin{eqnarray}%{rCl}
\bbeta^\top(I + \delta n^2\hat{\bSigma}_{\sfx\sfs}\hat{\bSigma}_{\sfs\sfx})\bbeta = \sum_{i=1}^{d}(1+\delta n^2\sigma_i^2)\beta_i^2 \nonumber
\end{eqnarray}
Since we aim at minimizing the penalty term $\bbeta^\top(I + \delta\hat{\bSigma}_{\sfx\sfs}\hat{\bSigma}_{\sfs\sfx})\bbeta$, the coefficient $\beta_i$ is likely to be low if the corresponding feature has high covariance with $\psi(\sfs)$, i.e. high $\sigma_i$. In the extreme case where $\delta \rightarrow \infty$, $\beta_i \rightarrow 0$ for all the features that have positive covariance with $\psi(\sfs)$. Moreover, if feature $i$ has $\sigma_i = 0$, its coefficient is unaffected by the extra penalization. In practice of course, $\hat{\bSigma}_{\sfs\sfx}$ is rarely diagonal, but the general idea is the same: the regularizer simply takes into account all cross-correlations to determine the penalty on each coefficient.

We now turn to the Bayesian perspective. Note that \eqref{fair_regression} is equivalent to the following Bayesian linear regression model
\begin{eqnarray}%{rCl}
\Y = \X\bbeta + \epsilon,& & \epsilon \sim  \cN(0, \lambda),\nonumber\\
\bbeta \sim  \cN(0, \bSigma),& &\bSigma = (\II + \delta n^2\hat{\bSigma}_{\sfx\sfs}\hat{\bSigma}_{\sfs\sfx})^{-1}.
\label{bayesian_linear}
\end{eqnarray}
Comparing to the normal Bayesian linear regression, we can see that this version simply modifies the prior on $\bbeta$. The same interpretation holds: assuming $\hat{\bSigma}_{\sfs\sfx}$ is diagonal and denoting the $i$-th diagonal element of $\hat{\bSigma}_{\sfx\sfs}\hat{\bSigma}_{\sfs\sfx}$ as $\sigma^2_i$, the prior covariance matrix $\bSigma$ is a diagonal matrix with $i$-th diagonal element of $(1+\delta n^2\sigma^2_i)^{-1}$. This means that we modify our prior such that the coefficients corresponding to the features with high $\sigma_i$ are shrank towards zero.%features with high $\sigma_i$ is more likely to be $0$. Since its prior variance is low, indicating a heavy probability mass around $0$. In the extreme case where $\eta_0 \rightarrow \infty$, we have $\Sigma_{ii} \rightarrow 0$ for all the features with $\sigma_i>0$. Furthermore, in the case $\lambda_i = 0$, the prior does not change.

\paragraph{Fair Kernel Ridge Regression} We now consider the nonlinear case with RKHSs $\mathcal{H}_k$ and $\mathcal{H}_{l}$ corresponding to feature maps $\phi(\cdot)$ and $\psi(\cdot)$ respectively. We denote the transformed data as $\bPhi$ and $\bPsi$ with the corresponding Gram matrices $\mathbf{K}$ and $\LL$. We extend the fair learning problem in the nonlinear case as the following optimization problem.
\begin{eqnarray}
\bar{\beta}:= \text{argmin} \frac{1}{n}\|\Y-\bPhi\bbeta\|_2^2 + \frac{\lambda}{n} \|\bbeta\|_2^2 + \eta \|\hat{\bSigma}_{\sfs\sfx}\bbeta\|_2^2 \nonumber\\
=\text{argmin} \frac{1}{\lambda}\|\Y-\bPhi\beta\|_2^2 + \bbeta^\top(\II + \delta n^2\hat{\bSigma}_{\sfx\sfs}\hat{\bSigma}_{\sfs\sfx})\bbeta. \label{fair_kernelregression} 
\end{eqnarray}
The interpretation of the form is similar to the linear case. We would like to penalize more for the coefficient $\beta_i$ if its corresponding feature has a large covariance with sensitive features $\sfs$.

Let us explore the Bayesian treatment of the nonlinear fair learning problem. In the weight space view, Eq.\eqref{fair_kernelregression} corresponds to the same model as \eqref{bayesian_linear}, with $\Y = \bPhi\bbeta + \epsilon$. However, we can readily derive the GP formulation. For any kernel $k$ where $k(\bfx,\bfx') = \langle \phi(\bfx), \phi(\bfx') \rangle$, the GP model is given by
\begin{eqnarray}%{rCl}
f \sim  \GP(0, k^*(\cdot,\cdot)),&&
y|f(\bfx) \sim  \cN(f(\bfx),\lambda),\nonumber\\
k^*(\bfx,\bfx') = \langle \phi(\bfx), \bSigma^* \phi(\bfx') \rangle,&&
\bSigma^* = (\II + \delta n^2 \hat{\bSigma}_{\sfx\sfs}\hat{\bSigma}_{\sfs\sfx})^{-1}.\label{GPRmodel}
\end{eqnarray}
%Since we do not have access to the true covariance operator, we rely on the empirical estimation, i.e. in practice, we would have $k^*(x,y) = \langle \phi(\bfx), \bSigma \phi(\bfy) \rangle$.
While it is not obvious that $k^*$ is tractable as it involves the operator $\Sigma^*:\mathcal H_k\to\mathcal H_k$, Proposition \ref{gp_conversion} proves that $k^*(\bfx,\bfx') = k(\bfx,\bfx') - k_{\bf{X}\bf{x}}^\top( \mathbf{K}\HH\LL\HH+\delta^{-1}\II)^{-1}\HH\LL\HH k_{\bf{X} \bf{x}'}$ and hence, one can readily employ this kernel as a modified GP prior and make use of the extensive GP modeling toolbox. We also note that we can treat kernel parameters of $k$ and $l$ as well as $\delta$ simply as parameters of $k^*$.

\section{Normalized dependence regularizers}
%\red{rewrite: Michael, Dino and Gustau. Should be a differnet section or a subsec?}

We have explained the fair kernel learning, and  introduced its corresponding Gaussian process version. Also, we provided another view of the dependence penalizer as the weighted norm of the coefficients where the weights are given by the cross-covariance operator. However, one issue with this framework is that the dependence measure is sensitive to the kernel parameters. For example, if we look at problem \eqref{fairlearning_hsic}, the extra penalty term $ \bff^\top \HH \LL \HH\bff $ is sensitive to the hyperparameters $\theta_k$ and $\theta_l$ from kernel $k$ and $l$. Notice that varying $\theta_l$ does not affect the other two terms in the objective function, one could simply adjust $\theta_l$ to reduce the HSIC value and hence reduce the objective function value. The unfairness however, is not reduced. Hence, one needs a parameter invariant dependence measure to avoid such an issue. As a result, we introduce the {\em normalized fair learning} framework in this section.

As shown in Eq.~\ref{fairlearning_hsic}, fairness is enforced through using HSIC value as the penalizer. Hence, a naive way of dealing with parameter sensitivity is to use the normalized version of HSIC. This has been extensively studied in \citep{Fukumizu:NIPS2007} where the so called NOCCO was proposed. Replacing HSIC with Hilbert-Schmidt norm of NOCCO in Eq.~\ref{fairlearning_hsic}, the fair learning is the following optimization problem:
\begin{equation}
    \min_{f\in\mathcal H_k} \bigg\{\frac{1}{n}\sum_{i=1}^n  V(f(x_i),y_i) + \lambda \Vert f \Vert _{\mathcal H_k}^2 +
    \eta \text{Tr}[\R_{f}\R_S]\bigg\},
    \label{fairlearning_nocco}
\end{equation}
where $\text{Tr}[\R_{f}\R_S]$ is the Hilbert-Schmidt norm of NOCCO between $f(\bfx)$ and ${\bf S}$. $\R_{f} = \HH\K\HH(\HH\K\HH+n\epsilon \II)^{-1}$, $\R_S = \HH\LL\HH(\HH\LL\HH+n\epsilon \II)^{-1}$ and $\epsilon$ is the regularization parameter used in the same way as in \citep{Fukumizu:NIPS2007}. Since we are using the linear kernel for $f(\bfx)$ and $\bff = \Phi \bbeta$, we have $\R_{f} = \HH\bPhi_x \bbeta \bbeta^\top\bPhi_x^\top\HH(\HH\bPhi_x \bbeta \bbeta^\top\bPhi_x^\top\HH+n\epsilon \II)^{-1}$. However, problem \eqref{fairlearning_nocco} does not admit a closed form solution. The reason is that the derivative of $\text{Tr}[\R_{f}\R_S]$ is not linear in $\bbeta$.
Hence, we ask the following question:

\begin{mdframed}[backgroundcolor=gray!20] \emph{Question 2}: can we find a normalized fair learning which admits a closed form solution?
\end{mdframed}

It turns out that the cross-covariance view of fair learning provides us a way to answer this question. In \eqref{fair_regression}, we used the empirical cross-covariance operator as the penalizer. To avoid the parameter sensitivity issue, we could use the normalized cross-covariance operator $\V_{\sfs\sfx}:= \bSigma_{\sfs\sfs}^{-1/2}\bSigma_{\sfs\sfx}\bSigma_{\sfx\sfx}^{-1/2}$ to replace $\bSigma_{\sfs\sfx}$. Let $\hat{\V}_{\sfs\sfx}:= \hat{\bSigma}_{\sfs\sfs}^{-1/2}\hat{\bSigma}_{\sfs\sfx}\hat{\bSigma}_{\sfx\sfx}^{-1/2}$ be the empirical version of $\V_{\sfs\sfx}$, the learning problem is now:
\begin{equation}
\bar{\bbeta} ~ := \text{argmin}_{\bbeta} \bigg\{\frac{1}{n}\|\bfy-\X\bbeta\|_2^2 + \lambda \|\bbeta\|_2^2 + \eta \|\hat{\V}_{\sfs\sfx}\bbeta\|_2^2\bigg\}.
\label{fair_regression_nocco} 
\end{equation}
This leads to a closed-form solution as
\begin{eqnarray}%{rCl}
\bar{\bbeta} &=& (\bPhi_x^\top\bPhi_x + n\lambda \II +n\eta\hat{\V}_{\sfx\sfs}\hat{\V}_{\sfs\sfx})^{-1}\bPhi_x^\top\bfy\nonumber\\
&=& (\bPhi_x^\top\bPhi_x + n\lambda I +n\eta\hat{\bSigma}_{\sfx\sfx}^{-1/2}\hat{\bSigma}_{\sfx\sfs}\hat{\bSigma}_{\sfs\sfs}^{-1}\hat{\bSigma}_{\sfs\sfx}\hat{\bSigma}_{\sfx\sfx}^{-1/2})^{-1}\bPhi_x^\top\bfy.\nonumber
\end{eqnarray}
In case where $\phi(\cdot)$ and $\psi(\cdot)$ are finite dimensional, the above provides a valid solution to the normalized fair learning problem. However, this is not the case if either of $\phi(\cdot)$ and $\psi(\cdot)$ is infinite dimensional. Since we face the problem of evaluating $\bSigma_{\sfx\sfx}^{-1/2}$ and $\bSigma_{\sfs\sfs}^{-1/2}$ terms which are infinite dimensional operators. 

To remedy this issue, we notice that the HSIC is potentially sensitive to parameters from $k$ and $l$. During the optimization process, parameters from $k$ is tuned from the data, while the parameters from $l$ are free to adjust. Hence, one could only partially normalize the cross-covariance operator with respect to hyperparameters from $l$ and formulate the following learning problem:
\begin{equation}
\bar{\beta} ~ := \text{argmin}\bigg\{ \frac{1}{n}\|\bfy-\X\bbeta\|_2^2 + \lambda \|\bbeta\|_2^2 \bigg\}
+ \eta \|\hat{\bSigma}_{\sfs\sfs}^{-1/2}\hat{\bSigma}_{\sfs\sfx}\bbeta\|_2^2
\label{fair_partial_nocco} 
\end{equation}
This gives us a closed-form solution as 
\begin{eqnarray}%{rCl}
\bar{\bbeta} &=& (\bPhi_x^\top\bPhi_x + n\lambda \II +n\eta\hat{\bSigma}_{\sfx\sfs}\hat{\bSigma}_{\sfs\sfs}^{-1}\hat{\bSigma}_{\sfs\sfx})^{-1}\bPhi_x^\top\bfy\nonumber\\
&=&\bPhi_x^\top(\K + n\lambda \II + \eta \K\tilde{\LL}(\tilde{\LL} + n\epsilon \II)^{-1})^{-1}\bfy\label{fair_nocco_beta}
\end{eqnarray}
where in the second equality we applied the Woodbury matrix inversion lemma\footnote{In computing $\hat{\bSigma}_{\sfs\sfs}^{-1}$, we use $(\hat{\bSigma}_{\sfs\sfs}+\epsilon \II)^{-1}$ instead to avoid the issue with non-invertible matrix.}.
As a result, the prediction at the training point for fair learning is \[\hat{\mathbf{f}}(\mathbf{x}) = \K(\K + n\lambda \II + \eta \K\tilde{\LL}(\tilde{\LL} + n\epsilon \II)^{-1})^{-1}\bfy.\]
\begin{remark}
We provide a justification for \eqref{fair_partial_nocco} via the conditional covariance operator. For any two random variable $\sfx$ and $\sfs$, we define the conditional covariance operator \[\bSigma_{\sfx\sfx|\sfs} = \bSigma_{\sfx\sfx}-\bSigma_{\sfx\sfs}\bSigma_{\sfs\sfs}^{-1}\bSigma_{\sfs\sfx}\footnote{For convenience, we have abused the notation $\bSigma_{\sfx\sfx}^{-1}$ since $\bSigma_{\sfs\sfs}$ can be non-invertible. But in this case, we can always use the regularized version $(\bSigma_{\sfx\sfx} + \epsilon \II)^{-1}$ instead. }. \] It has been shown in \cite[Proposition 2]{fukumizu2009kernel} that 
\begin{eqnarray}
\langle \bbeta, \bSigma_{\sfx\sfx|\sfs}\bbeta \rangle &=& \langle \bbeta, \bSigma_{\sfx\sfx}\bbeta \rangle -\langle \bbeta, \bSigma_{\sfx\sfs}\bSigma_{\sfs\sfs}^{-1}\bSigma_{\sfs\sfx}\bbeta \rangle \nonumber\\
&=& \inf_{g\in \mathcal{H}_l}\mathbb{E}_{\sfx \sfs}|f(x) - g(s)|^2\label{con_cov1}
\end{eqnarray}
Notice that Eq.\eqref{con_cov1} is the minimal residual error when we use $g(s)$ to predict $f(x)$, for any $g \in \mathcal{H}_l$. In other words, it is the variance in $f(x)$ that cannot be explained by $g(s)$. Since $\langle \bbeta, \bSigma_{\sfx\sfx}\bbeta \rangle$ represents the variance of $f(x)$, we can treat $\langle \bbeta, \bSigma_{\sfx\sfs}\bSigma_{\sfs\sfs}^{-1}\bSigma_{\sfs\sfx}\bbeta \rangle$ as the maximal amount of variance of $f(x)$ that can be explained by $g(s)$. In \eqref{fair_partial_nocco}, \[\|\hat{\bSigma}_{\sfs\sfs}^{-1/2}\hat{\bSigma}_{\sfs\sfx}\bbeta\|_2^2  =\langle \bbeta, \hat{\bSigma}_{\sfx\sfs}\hat{\bSigma}_{\sfs\sfs}^{-1}\hat{\bSigma}_{\sfs\sfx}\bbeta \rangle,\]
minimizing this term is equivalent to minimize the amount of variance in $f(x)$ that can be explained by $g(s), \forall g \in \mathcal{H}_l$. This is essentially the same as minimizing the dependence between $f(x)$ and $\sfs$. Furthermore, if $l$ is a universal kernel ( e.g. Gaussian kernel, Laplace kernel, etc., refer to \citep{sriperumbudur2011universality} for more details on universal kernel), Eq.\eqref{con_cov1} can be rewritten as
\begin{eqnarray}
\langle \bbeta, \bSigma_{\sfx\sfx|\sfs}\bbeta \rangle= \inf_{g\in L^2}\mathbb{E}_{\sfx \sfs}|f(x) - g(s)|^2,\label{con_cov2}
\end{eqnarray}
where $L^2$ is the space of all square integrable functions defined on $\mathcal S$. In this case, $\langle \bbeta, \bSigma_{\sfx\sfs}\bSigma_{\sfs\sfs}^{-1}\bSigma_{\sfs\sfx}\bbeta \rangle$ quantifies the maximal amount of variance in $f(x)$ that can be explained by $g(s), \forall g \in L^2$. Note that $L^2$ is independent of the choice of $l$, this is particularly useful in the normalized fair learning problem. The reason is, although in defining $\bSigma_{\sfx\sfs}$ we rely on the hyperparameter $\theta_l$ from kernel $l$, the quantity $\langle \bbeta, \bSigma_{\sfx\sfs}\bSigma_{\sfs\sfs}^{-1}\bSigma_{\sfs\sfx}\bbeta \rangle$ is independent of $l$. In other words, varying $\theta_l$ will not affect its value. This justifies the usage of $\|\hat{\bSigma}_{\sfs\sfs}^{-1/2}\hat{\bSigma}_{\sfs\sfx}\bbeta\|_2^2$ as the penalty term in normalized fair learning.
\end{remark}

\section{Experiments}

In this section, we illustrate the performance of the proposed methods on both synthetic and real-data problems, and study the effect of the fairness regularization. We first study performance in simulated toy datasets that allow us to study the error-vs-dependence paths and demonstrate the potential of proposed approaches in controlled scenarios. Secondly, we study the effect of the normalized dependence regularizer as well as the use of the GP formulation in contrast to the ERM framework, i.e. kernel ridge regression, in two real-data fairness problems: crime prediction and income prediction. %Finally, we give empirical evidence of performance in cases of enforcing the GP model to be physically consistent with either physical models or ancillary data, and how can the dependence term plays the role of explaining the plausability of data and models. %illustrative examples involving GP models that are required to be faithful with respect to physical models or ancillary data, as well as GP models that try to accommodate consistency with climate change attribution.

\subsection{Toy dataset 1}

We start by demonstrating the effectiveness of the proposed fairness framework by comparing it to two other baselines based on the fairness literature. The first approach is simple omission of the sensitive variable (OSV), where we use all the features except the prespecified sensitive variable. The second one mimics the ideas of fair representation learning (FRL) \cite{ZemelICML13} where the input data is transformed such that it contains as much information as possible from the original data while simultaneously being statistically independent from the sensitive variable. The transformed data is then used for learning. The dataset we consider is as follows: we first sample $x_1,x_2,z$ independently from $\mathcal{N}(0,1)$; assuming $z$ is unobserved, we let the sensitive variable be $x_3 = \frac{1}{\sqrt{2}}(x_1 + z)$. Obviously, $x_1$ and $x_3$ are correlated. Let the true function of interest be \[f(\mathbf{x},z) = \text{sign}((x_1-z)x_3)|x_2|,\] where $\mathbf{x} = [x_1,x_2,x_3]^T$. It is readily checked that $f(\mathbf{x},z)$ is marginally independent of the sensitive variable $x_3$. We now further assume that the observations $y$ include a bias that is based on the sensitive variable $x_3$ 
\[y = f(\mathbf{x},z) + 2b\mathbf{1}_{\{x_3>0\}}-b + \epsilon,\]
i.e. the observations are on average increased by $b$ when $x_3>0$ and decreased by $b$ otherwise.
Given data $\{\mathbf{x}_i,y_i\}_{i}^n$, our task is to find a best fit while preserving fairness in terms of statistical parity. Clearly, simply removing the sensitive variable while training the model is not appropriate as the bias in the observations is correlated with $x_1$ as well and will thus be retained. Alternatively, we may want to fully remove all dependence on $x_3$ from the inputs. This simply corresponds to transforming $x_1$ as follows:
\begin{equation}
    \tilde{x}_{1}=x_{1}-\frac{\mathbb{E}\left[x_{1}x_{3}\right]}{\mathbb{E}x_{3}^{2}}x_{3}=x_{1}-\frac{1}{\sqrt{2}}x_{3}=\frac{1}{2}\left(x_{1}-z\right).
\end{equation}
However, this shows the danger of such an approach -- we now have input $\tilde x_1$ independent of the sensitive variable, but the true function $f(\mathbf{x},z)$ is marginally independent of $\tilde x_1$ as well and hence the transformed variable will not be useful for learning! Hence, the fairness regularization on the predictor provides a remedy -- it directly penalizes the dependence between the predictor and the sensitive variable rather than between the inputs and the sensitive variable, which does not take into account the learning problem at hand. We compare the performances between the following approaches: standard kernel ridge regression (KRR) and Gaussian process regression (GPR) without data modification, fairness regularization (both KRR and GPR versions) with different $\eta$ (refer to Eq. \ref{fairlearning_hsic}) values; OSV and FRL. In the case of kernel ridge regression (KRR), we choose the kernel lengthscale and regularization parameter with cross-validation and in the GP versions, we choose them via maximization of the marginal likelihood. We measure the performance of each model through the coefficient of determination $R^2 = \frac{\text{variance explained by the predictor}}{\text{total variance}}$ with respect to both the observed responses $y_i$ and the true function values. By definition of $R^2$, we would expect the standard approach to achieve the highest $R^2$ (on biased data) as it utilizes all the available information, whereas FRL would have the lowest score. For fairness regularization, this will depend on the value of $\eta$, i.e. model with high $\eta$ will have low $R^2$. Looking at Table \ref{table_r2}, we do see this pattern. On the other hand, if we consider $R^2$ on the true function values, we see that it tends to increase with higher $\eta$, i.e. higher fairness regularization improves the removal of the bias present in the observed responses from the predictors. As expected, FRL detects no signal on this data, and OSV also leads to a significant drop in $R^2$. In addition, since the GP version allows us to systematically select the hyperparameters, we can see that in most cases, $R^2$ from the GP model will be higher than its kernel regression version. We next report the correlation between the predicted value $\hat{y}$ and $x_3$. Likewise, we would expect the standard approach will have the highest correlation while the FRL will have the lowest correlation. For fairness regularization, the correlation decreases as $\eta$ increases. Table \ref{table_coef} reports these results. We see that OSV still has a high correlation to the sensitive variable $x_3$. In contrast, the GPR for $\eta=200$ allows a predictor that is essentially uncorrelated from $x_3$ while having strong $R^2$ performance.

\begin{table}[h!]
\small
\centering
\def\arraystretch{.75}
\caption{The $R^2$ wrt. observations (left) and wrt. true value (right).}\label{table_r2}
\begin{tabular}{ccc|cc}
\hline\hline
Approach      & KRR &  GPR  & KRR & GPR\\[0.5ex]
\hline
\hline
Standard & 0.606 $\pm$ 0.002    & 0.612 $\pm$ 0.002 & 0.332 $\pm$ 0.003 & 0.356 $\pm$ 0.003 \\
\hline
$\eta = 2\times 10^{-3}$ & 0.600 $\pm$ 0.002    & 0.610 $\pm$ 0.001 & 0.358 $\pm$ 0.003 &  0.335 $\pm$ 0.002 \\
$\eta = 0.2$ & 0.567 $\pm$ 0.001    & 0.586 $\pm$ 0.009  & 0.341 $\pm$ 0.005 & 0.394 $\pm$ 0.010\\
$\eta = 20$ &  0.488 $\pm$ 0.008   &  0.506 $\pm$ 0.012  & 0.466 $\pm$ 0.011 & 0.472 $\pm$ 0.008\\
$\eta = 200$ & 0.384 $\pm$ 0.011 &  0.403 $\pm$ 0.005 & 0.321 $\pm$ 0.014& 0.530 $\pm$ 0.004 \\
\hline
OSV & 0.238 $\pm$ 0.007     & 0.196 $\pm$ 0.013   & 0.123 $\pm$ 0.008 & 0.098 $\pm$ 0.019\\
\hline
FRL & -0.021 $\pm$ 0.002 & -0.009 $\pm$ 0.001 &-0.024 $\pm$ 0.002 &  -0.011 $\pm$ 0.001\\
\hline
\hline
\end{tabular}
\label{table:disrmse1}
\end{table}

\begin{table}[h!]
\small
\centering
\def\arraystretch{.75}
\caption{The correlation between $\hat{y}$ and $x_3$.}\label{table_coef}
\begin{tabular}{ccc}
\hline\hline
Approach      & KRR &  GPR \\[0.5ex]
\hline
\hline
Standard & 0.3917 $\pm$ 0.0011    & 0.3863 $\pm$ 0.0013   \\
\hline
$\eta = 2\times 10^{-3}$ & 0.4053 $\pm$ 0.0019    & 0.3853 $\pm$ 0.0024    \\
$\eta = 0.2$ & 0.3337 $\pm$ 0.0104    & 0.3257 $\pm$ 0.0206   \\
$\eta = 20$ &  0.1364 $\pm$ 0.0150     & 0.2234 $\pm$ 0.0455   \\
$\eta = 200$ & 0.1066 $\pm$ 0.0078 &  0.0139 $\pm$ 0.0031\\
\hline
OSV & 0.2976 $\pm$ 0.0053    & 0.3195 $\pm$ 0.0058    \\
\hline
FRL & -0.0010 $\pm$ 0.0012 & -0.0102 $\pm$ 0.0013\\
\hline
\hline
\end{tabular}
\label{table:disrmse2}
\end{table}

\subsection{Toy dataset 2} 

We next consider a simple simulated dataset following the model from \cite{perez2017fair}:
\begin{eqnarray}%{rCl}
{\sfy} = {\sfx}^2 + {\sfs}^2 + \epsilon ,&& {\sfx}|\sfs \sim \mathcal{N} (\log(|\sfs|),\sigma_{\sfx}^2),\nonumber\\
{\sfs} \sim   \mathcal{N}(0,\sigma_{\sfs}^2),&& \epsilon \sim  \mathcal{N} (0,\sigma_\sfy^2).\nonumber
\end{eqnarray}

\begin{figure}[t!]
\includegraphics[width=.47\columnwidth]{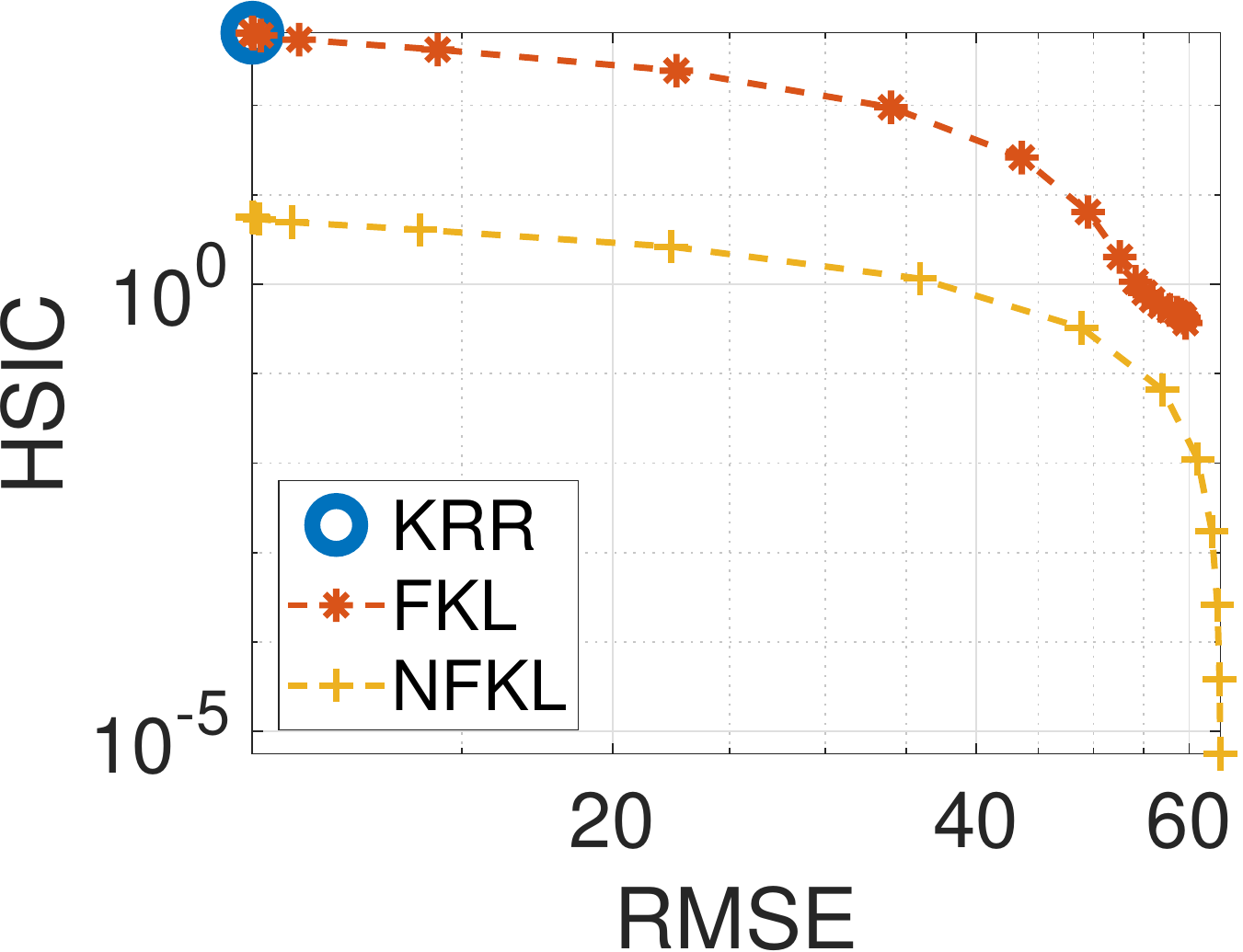}
\includegraphics[width=.47\columnwidth]{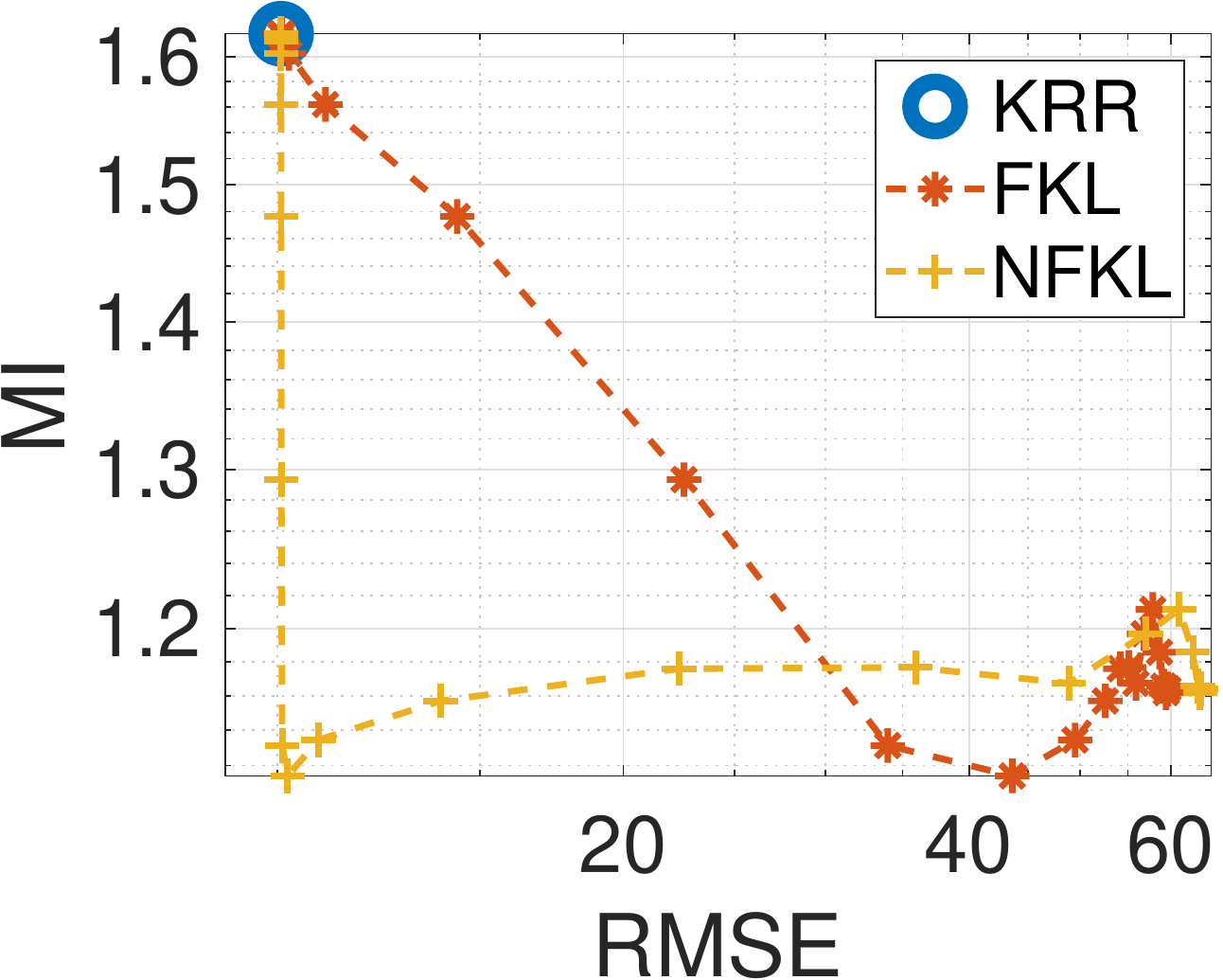}
\caption{(left) RMSE vs. HSIC of the ERM versions for different values of the fairness parameter. (right) RMSE vs. Mutual Information of the ERM versions for different values of the fairness parameter.}\label{exp:synthetic}
\end{figure}

Similarly as in the previous example, even if we omit the sensitive variable ${s}$, the remaining variables are dependent on it. %the sensitive variable.
We will use this dataset to study the impact of normalizing the HSIC regularizer on the trade-offs between the predictive performance and dependence on the sensitive variable. 
%\paragraph{The effect of normalization}
We compare here the following methods: Kernel Ridge Regression (KRR), Fair Kernel Learning (FKL), and the Normalized Fair Kernel Learning (NFKL) on toy dataset 2. To validate the behavior of the proposed methods we used the RMSE as an error measurement of the predictions. As a fairness measurement we used both the HSIC and Mutual Information (MI) estimates between the output predictions and the sensitive variables. We performed $50$ trials using $n=700$ points for training algorithms and $700$ points for the final test validation. We chose $25$ different values for the fairness parameter $\eta$ logarithmically spaced in the range $[10^{-7},10^3]$. In the case of the kernel lengthscale and regularization parameters we did cross-validation taking $10$ values logarithmically spaced in ranges $\theta\in[10^{-4},10^3]$ and $\lambda\in[10^{-4},10^4]$. In the case of NFKL we have fixed the parameter $\epsilon=10^{-6}$. Figure~\ref{exp:synthetic} illustrates the averaged results of the presented methods. The standard KRR method (corresponding to case $\eta=0$) achieves the best performance in RMSE, but it is the also the most unfair in terms of both dependence measures. The use of the proposed fairness regularization approaches is able to mitigate the unfairness of the predictors by trading it off for the RMSE as the fairness regularization parameter $\eta$ is varied producing the unfairness/error curves shown in Figure ~\ref{exp:synthetic}. We see that NFKL outperforms FKL, i.e. that the normalization of the regularizer substantially improves this tradeoff.

\subsection{Crime and income prediction} 

In the next set of experiments, we empirically compare the performance of fair kernel learning and our proposed GP version on two real datasets: 
\begin{itemize}
    \item {\em Communities and Crime} \citep{redmond2002data}.
    %For the Communities and Crime dataset, 
    We are here concerned about predicting per capita violent crime rate in different communities in the United States from a set of relevant features, such as median family income or the percentage of people under poverty line. Race is considered the sensitive variable. The dataset contains $1994$ instances with $127$ features. Some of the features contained many missing values as some surveys were not conducted in some communities, so they were removed from the data. % We also deleted one instance that has two many missing values. 
%After this operation, 
This returns a $1993 \times 100$ data matrix. We will use this data to assess performance of the discriminative versus the GP-based algorithm.
%In both simulated data and real data cases,
\item {\em Adult Income} \citep{Dua:2017}. 
The Adult dataset contains $48841$ subjects, which consists of $32561$ training data and $16581$ data. The original data have $14$ features among which $6$ are continuous and the remaining are categorical. The label is binary indicating whether a subjects's income is higher that $50$K or not. Each continuous feature was then discretized into quantiles and represented by a binary variable. Hence the final dataset has 123 features. %We would like to 
The goal is to predict a given subject's income level while controlling for the sensitive variables: gender and race. We preprocessed the data so that each feature of the predictor variable $\sfx$ as well as the response variable $\sfy$ has zero mean and standard deviation $1$.
\end{itemize}

\paragraph{Fair KRR vs Fair GP}

We empirically compared the performance of the two fair kernel learning model: kernel ridge regression and the modified GP version. In the regression setting, we used $5$-fold cross validation to choose the kernel bandwidth parameter $\theta_k$ for $k$ and the penalty parameter $\lambda$. For the sake of a fair comparison, we have set the parameter for the kernel on the sensitive variable $s$ to be fixed at $\theta_l = 0.5$, while we select $\theta_k$ according to the median heuristic and also randomly draw $10$ samples around its value. In addition, we draw $15$ values between $[e^{-15}, 1.0]$.

\begin{figure}[h!]%\label{krr_and_gp}
\centerline{\includegraphics[width=1.1\columnwidth]{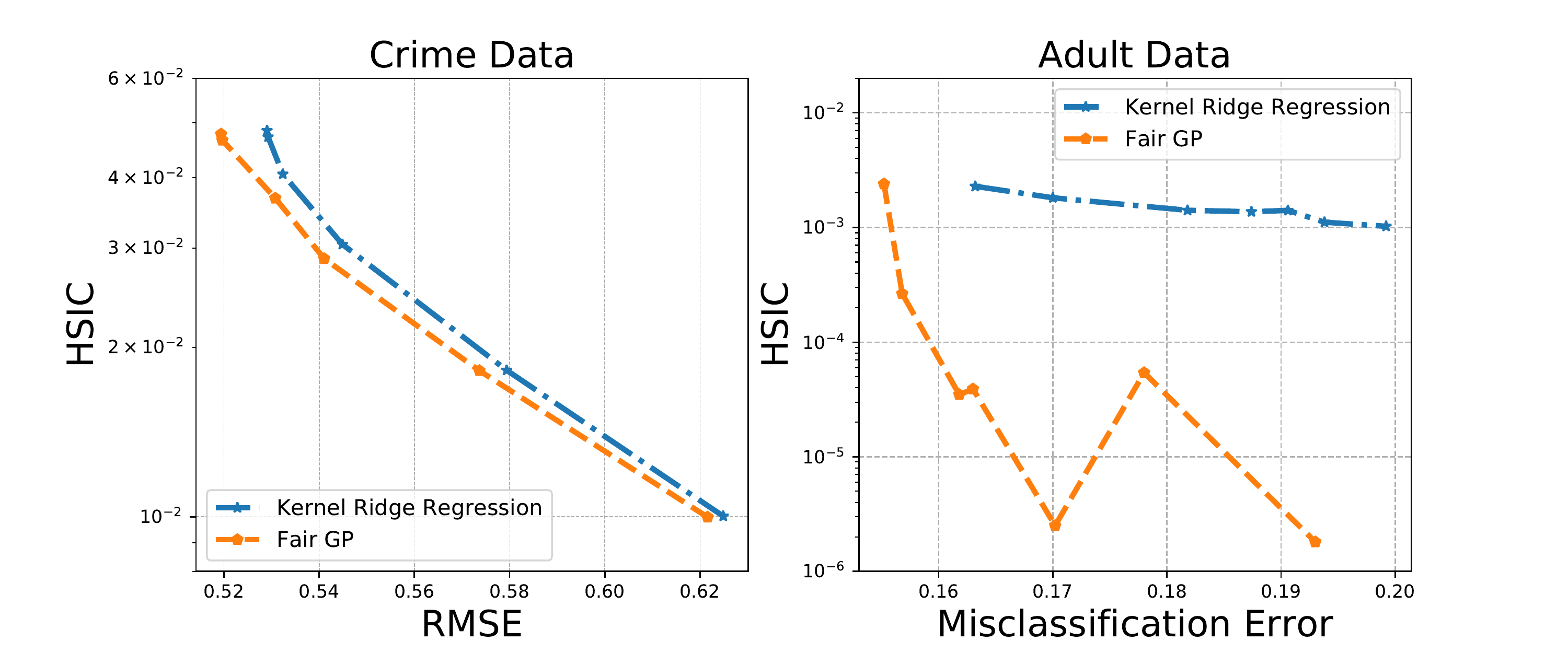}}
\caption{Comparisons between fairness-regularized kernel ridge regression and fairness-regularized GP regression on crime and income data. Principled hyperparameter selection due to the proposed GP model allows improved unfairness / prediction error trade-off curves.} %\textcolor{red}{this figure needs to be improved to look like Fig 1: thicker lines, log-scales, larger fonts, background grid....}}
\label{krr_and_gp}
\end{figure}

For the modified GP case, we fixed $\theta_l = 0.5$ as in the regression setting while optimizing over other hyperparameters. In both settings, we chose $7$ different $\eta$ (the penalty hyperparameter for unfairness) in the interval $[0,10]$ with high $\eta$ value representing more fair model. Figure \ref{krr_and_gp} demonstrate the result for the two model in %simulation data and 
the crime data. We can see that for most of the time, the modified GP outperforms kernel ridge regression. The modified GP gives better trade-off between fairness and prediction accuracy due to its optimization process. Note that the performance gap between fair kernel learning and fair GP is larger in the Adult data than that in the Crime data. An possible reason is that the Crime data is much harder to learn (RMSE is 0.6 at its highest) so that the advantage of using GP in optimizing hyperparameter is limited.

\paragraph{Fair GP with ARD Kernel}

In this section, we show empirical evidence of the performance of the modified GP fair learning with ARD (Automatic Relevance Determination) kernel to the Communities and Crime real dataset. The goal of this experiment setting is to assess the effect of the fair learning on the coefficients for each feature. Specifically, we run two sets of experiments. The first experiment is to perform predictions with standard GP. The kernel is set to be ARD RBF defined as :
\[\text{for} ~\bfx,\bfx' \in \mathbb{R}^d, ~k(\bfx,\bfx') = \exp\bigg(-\sum_{i=1}^d \theta_i^{-2}(x_i-x'_i)^2\bigg).\]

This experiment is similar to the previous one, except that we used the modified GP framework. We would like to see how the $\theta_i$'s for those sensitive variables change when we impose the fairness regularizer. We list the results in Table \ref{table_coefficients}. The root mean square error and the unfairness of the predictions in two settings are also reported in the table.

%\red{GCV: I cannot find the data description and results for the adult dataset, this part is weak and only contains fig2b, right? For the description, you can copy from https://arxiv.org/abs/1710.05578 Did you manage with the results with the ARD kernel?}

\begin{table}[h!]
\small
\centering
\def\arraystretch{.75}
\caption{The change of $\theta_i$ for sensitive variables with and without fair lerning}\label{table_coefficients}
\begin{tabular}{lll}
\hline\hline
Sensitive Variable      & GP &  Fair GP \\[0.5ex]
\hline
Race-Black & 1.809 $\pm$ 0.216     & 2.939 $\pm$ 0.367    \\
Race-White & 6.728 $\pm$ 3.425     & 2.519 $\pm$ 0.038    \\
Race-Asian & 17.79 $\pm$ 11.96    & 117.9 $\pm$ 0.045    \\
Race-Hispanic & 53.90 $\pm$ 19.00    & 9.669 $\pm$ 1.606   \\

Income-White & 132.2 $\pm$ 2.823     & 213.0 $\pm$ 0.190   \\
Income-Black & 108.9 $\pm$ 88.73    & 389.3 $\pm$ 0.026    \\
Income-Indian & 176.4 $\pm$ 7.351     & 700.9 $\pm$ 0.014    \\
Income-Asian& 17.76 $\pm$ 8.051     & 386.2 $\pm$ 0.077   \\
Income-Other &  12.63 $\pm$ 6.762     & 411.3 $\pm$ 0.136   \\
Income-Hispanic &175.2 $\pm$ 4.667  &404.7 $\pm$ 0.020 \\
RMSE & 0.627 $\pm$ 0.054  &0.766 $\pm$ 0.036\\
Unfairness &0.050 $\pm$ 0.001  &0.0024 $\pm$ 0.0001 \\
\hline
\end{tabular}
\label{table:disrmse3}
\end{table}

We can see that for most sensitive variables, their bandwidths were significantly increased after performing fair learning. This means that in computing the kernel value, those sensitive variables are contributing less, i.e. we treat instances as similar even when their sensitive variables have different values and as a result, the learned function varies less in those dimensions than in others.

\section{Conclusions}

% GCV: let's call it a thing
Using machine learning to facilitate and automate data-informed decisions has a huge potential to benefit society and transform people's lives. However, data used to train machine learning models are not necessarily free from cognitive or other biases, so the discovered patterns may retain or compound discriminatory decisions. We introduced a regularization framework of fairness-aware models where statistical dependence between predictions and the sensitive, protected variables is penalized. The use of kernel dependence measures as fairness regularizers allowed us to obtain simple regression models with closed-form solutions, derive a probabilistic Gaussian process interpretation, as well as the appropriate normalization of the regularizers. The latter two developments lead to principled and robust hyperparameter selection. The developed methods show promising performance in synthetic and real-data experiments involving crime and income prediction, allowing to strike favourable tradeoffs between method's predictive performance (on biased data) and its fairness in terms of statistical parity. While we focused on a specific viewpoint on fairness here, considering directly the statistical dependence on a prespecified set of sensitive variables, construction of machine learning techniques suited for other notions of fairness involving causal associations and conditional dependencies presents an important future research challenge. As there is also a flurry of research on the use of kernel methods in these fields, similar approaches invoking appropriate notions of kernel-based regularizers may be possible.

\section{Acknowledgements}
A.P.-S. and G.C.-V. are supported by the European Research Council (ERC) under the ERC-CoG-2014 SEDAL Consolidator grant (grant agreement 647423). D.S. is supported in part by The Alan Turing Institute (EP/N510129/1). The authors thank Kenji Fukumizu for fruitful discussions and Alan Chau, Lucian Chan, Qinyi Zhang and Alex Shestopaloff for helpful comments.

%under \red{gender and race} discrimination problems. The GP treatment also permitted studying the relative relevance of covariates under fairness constraints, \red{as well as quantifying population and individual biases.}
%\red{some final words about the future work are needed here, and to end with a nice statement of the opportunities the framework may offer...}

% {\small
% \itemsep
\bibliographystyle{elsarticle-harv} 
\bibliography{references.bib}

\begin{thebibliography}{38}
\expandafter\ifx\csname natexlab\endcsname\relax\def\natexlab#1{#1}\fi
\expandafter\ifx\csname url\endcsname\relax
  \def\url#1{\texttt{#1}}\fi
\expandafter\ifx\csname urlprefix\endcsname\relax\def\urlprefix{URL }\fi

\bibitem[{Adebayo and Kagal(2016)}]{adebayo2016iterative}
Adebayo, J., Kagal, L., 2016. Iterative orthogonal feature projection for
  diagnosing bias in black-box models. arXiv preprint arXiv:1611.04967.

\bibitem[{Belkin et~al.(2006)Belkin, Niyogi, and Sindhwani}]{Belkin2006}
Belkin, M., Niyogi, P., Sindhwani, V., Dec. 2006. Manifold regularization: A
  geometric framework for learning from labeled and unlabeled examples. Journal
  of Machine Learning Research 7, 2399--2434.

\bibitem[{Berk et~al.(2018)Berk, Heidari, Jabbari, Kearns, and
  Roth}]{berk2018fairness}
Berk, R., Heidari, H., Jabbari, S., Kearns, M., Roth, A., 2018. Fairness in
  criminal justice risk assessments: The state of the art. Sociological Methods
  \& Research, 0049124118782533.

\bibitem[{Brennan and Ehret(2009)}]{brennan2009evaluating}
Brennan, Tim, W.~D., Ehret, B., 2009. Evaluating the predictive validity of the
  compas risk and needs assessment system. Criminal Justice and Beh. 36~(1),
  21--40.

\bibitem[{Calders and Verwer(2010)}]{calders2010three}
Calders, T., Verwer, S., 2010. Three naive bayes approaches for
  discrimination-free classification. Data Mining and Knowledge Discovery
  21~(2), 277--292.

\bibitem[{Calmon et~al.(2017)Calmon, Wei, Vinzamuri, Ramamurthy, and
  Varshney}]{calmon2017optimized}
Calmon, F., Wei, D., Vinzamuri, B., Ramamurthy, K.~N., Varshney, K.~R., 2017.
  Optimized pre-processing for discrimination prevention. In: Advances in
  Neural Information Processing Systems. pp. 3992--4001.

\bibitem[{Chouldechova(2017)}]{Chouldechova17}
Chouldechova, A., 2017. Fair prediction with disparate impact: A study of bias
  in recidivism prediction instruments. Big data 5~(2), 153--163.

\bibitem[{Chouldechova and Roth(2018)}]{chouldechova2018frontiers}
Chouldechova, A., Roth, A., 2018. The frontiers of fairness in machine
  learning. arXiv preprint arXiv:1810.08810.

\bibitem[{Cunningham and Sorensen(2006)}]{cunningham2006actuarial}
Cunningham, M.~D., Sorensen, J.~R., 2006. Actuarial models for assessing prison
  violence risk: revisions and extensions of the risk assessment scale for
  prison (rasp). Assessment 13~(3), 253--265.

\bibitem[{Dheeru and Karra~Taniskidou(2017)}]{Dua:2017}
Dheeru, D., Karra~Taniskidou, E., 2017. {UCI} machine learning repository.
\newline\urlprefix\url{http://archive.ics.uci.edu/ml}

\bibitem[{Dieterich et~al.(2016)Dieterich, Mendoza, and
  Brennan}]{dieterich2016compas}
Dieterich, W., Mendoza, C., Brennan, T., 2016. Compas risk scales:
  Demonstrating accuracy equity and predictive parity. Northpoint Inc.

\bibitem[{Dwork et~al.(2012)Dwork, Hardt, Pitassi, Reingold, and
  Zemel}]{dwork2012fairness}
Dwork, C., Hardt, M., Pitassi, T., Reingold, O., Zemel, R., 2012. Fairness
  through awareness. In: Proceedings of the 3rd innovations in theoretical
  computer science conference. ACM, pp. 214--226.

\bibitem[{Feldman et~al.(2015)Feldman, Friedler, Moeller, Scheidegger, and
  Venkatasubramanian}]{feldman2015certifying}
Feldman, M., Friedler, S.~A., Moeller, J., Scheidegger, C., Venkatasubramanian,
  S., 2015. Certifying and removing disparate impact. In: Proceedings of the
  21th ACM SIGKDD International Conference on Knowledge Discovery and Data
  Mining. pp. 259--268.

\bibitem[{Fukumizu et~al.(2009)Fukumizu, Bach, Jordan,
  et~al.}]{fukumizu2009kernel}
Fukumizu, K., Bach, F.~R., Jordan, M.~I., et~al., 2009. Kernel dimension
  reduction in regression. The Annals of Statistics 37~(4), 1871--1905.

\bibitem[{Fukumizu et~al.(2008)Fukumizu, Gretton, Sun, and
  Sch\"{o}lkopf}]{Fukumizu:NIPS2007}
Fukumizu, K., Gretton, A., Sun, X., Sch\"{o}lkopf, P.~B., 2008. Kernel measures
  of conditional dependence. In: Advances in Neural Information Processing
  Systems 20. pp. 489--496.

\bibitem[{Gretton et~al.(2012)Gretton, Borgwardt, Rasch, Sch\"{o}lkopf, and
  Smola}]{Gretton2012}
Gretton, A., Borgwardt, K.~M., Rasch, M.~J., Sch\"{o}lkopf, B., Smola, A., Mar.
  2012. A kernel two-sample test. J. Mach. Learn. Res. 13, 723--773.

\bibitem[{Gretton et~al.(2005)Gretton, Herbrich, and Hyv{\"a}rinen}]{Gretton05}
Gretton, A., Herbrich, R., Hyv{\"a}rinen, A., 2005. Kernel methods for
  measuring independence. Journal of Machine Learning Research 6, 2075--2129.

\bibitem[{Hardt et~al.(2016)Hardt, Price, Srebro, et~al.}]{hardt2016equality}
Hardt, M., Price, E., Srebro, N., et~al., 2016. Equality of opportunity in
  supervised learning. In: Advances in neural information processing systems.
  pp. 3315--3323.

\bibitem[{Heidari et~al.(2018)Heidari, Ferrari, Gummadi, and
  Krause}]{heidari2018fairness}
Heidari, H., Ferrari, C., Gummadi, K., Krause, A., 2018. Fairness behind a veil
  of ignorance: A welfare analysis for automated decision making. In: Advances
  in Neural Information Processing Systems. pp. 1265--1276.

\bibitem[{Hoffman et~al.(2017)Hoffman, Kahn, and Li}]{hoffman2017discretion}
Hoffman, M., Kahn, L.~B., Li, D., 2017. Discretion in hiring. The Quarterly
  Journal of Economics 133~(2), 765--800.

\bibitem[{Joseph et~al.(2016)Joseph, Kearns, Morgenstern, and
  Roth}]{joseph2016fairness}
Joseph, M., Kearns, M., Morgenstern, J.~H., Roth, A., 2016. Fairness in
  learning: Classic and contextual bandits. In: Advances in Neural Information
  Processing Systems. pp. 325--333.

\bibitem[{Kamiran and Calders(2009)}]{kamiran2009classifying}
Kamiran, F., Calders, T., 2009. Classifying without discriminating. In: 2009
  2nd International Conference on Computer, Control and Communication. IEEE,
  pp. 1--6.

\bibitem[{Kamiran and Calders(2012)}]{kamiran2012data}
Kamiran, F., Calders, T., 2012. Data preprocessing techniques for
  classification without discrimination. Knowledge and Information Systems
  33~(1), 1--33.

\bibitem[{Kamishima et~al.(2012)Kamishima, Akaho, Asoh, and
  Sakuma}]{kamishima2012fairness}
Kamishima, T., Akaho, S., Asoh, H., Sakuma, J., 2012. Fairness-aware classifier
  with prejudice remover regularizer. In: Joint European Conference on Machine
  Learning and Knowledge Discovery in Databases. Springer, pp. 35--50.

\bibitem[{Kanagawa et~al.(2018)Kanagawa, Hennig, Sejdinovic, and
  Sriperumbudur}]{kanagawa2018gaussian}
Kanagawa, M., Hennig, P., Sejdinovic, D., Sriperumbudur, B.~K., 2018. Gaussian
  processes and kernel methods: A review on connections and equivalences. arXiv
  preprint arXiv:1807.02582.

\bibitem[{Kim et~al.(2018)Kim, Reingold, and Rothblum}]{kim2018fairness}
Kim, M., Reingold, O., Rothblum, G., 2018. Fairness through
  computationally-bounded awareness. In: Advances in Neural Information
  Processing Systems. pp. 4842--4852.

\bibitem[{Kimeldorf and Wahba(1970)}]{kimeldorf1970}
Kimeldorf, G.~S., Wahba, G., 1970. A correspondence between {B}ayesian
  estimation on stochastic processes and smoothing by splines. Ann. Math.
  Statist. 41~(2), 495--502.

\bibitem[{Kleinberg et~al.(2016)Kleinberg, Mullainathan, and
  Raghavan}]{kleinberg2016inherent}
Kleinberg, J., Mullainathan, S., Raghavan, M., 2016. Inherent trade-offs in the
  fair determination of risk scores. arXiv preprint arXiv:1609.05807.

\bibitem[{Luo et~al.(2015)Luo, Liu, Koprinska, and
  Chen}]{luo2015discrimination}
Luo, L., Liu, W., Koprinska, I., Chen, F., 2015. Discrimination-aware
  association rule mining for unbiased data analytics. In: International
  Conference on Big Data Analytics and Knowledge Discovery. Springer, pp.
  108--120.

\bibitem[{Pedreschi et~al.(2008)Pedreschi, Ruggieri, and
  Turini}]{Pedreschi:2008}
Pedreschi, D., Ruggieri, S., Turini, F., 2008. Discrimination-aware data
  mining. In: Proceedings of the 14th ACM SIGKDD International Conference on
  Knowledge Discovery and Data Mining. KDD '08. pp. 560--568.

\bibitem[{P{\'e}rez-Suay et~al.(2017)P{\'e}rez-Suay, Laparra,
  Mateo-Garc{\'\i}a, Mu{\~n}oz-Mar{\'\i}, G{\'o}mez-Chova, and
  Camps-Valls}]{perez2017fair}
P{\'e}rez-Suay, A., Laparra, V., Mateo-Garc{\'\i}a, G., Mu{\~n}oz-Mar{\'\i},
  J., G{\'o}mez-Chova, L., Camps-Valls, G., 2017. Fair kernel learning. In:
  Joint European Conference on Machine Learning and Knowledge Discovery in
  Databases. Springer, pp. 339--355.

\bibitem[{Redmond and Baveja(2002)}]{redmond2002data}
Redmond, M., Baveja, A., 2002. A data-driven software tool for enabling
  cooperative information sharing among police departments. European Journal of
  Operational Research 141~(3), 660--678.

\bibitem[{Ristanoski et~al.(2013)Ristanoski, Liu, and Bailey}]{Ristanoski2013}
Ristanoski, G., Liu, W., Bailey, J., 2013. Discrimination aware classification
  for imbalanced datasets. In: CIKM '13. ACM, NY, USA, pp. 1529--1532.

\bibitem[{Ruggieri et~al.(2010)Ruggieri, Pedreschi, and Turini}]{Ruggieri:2010}
Ruggieri, S., Pedreschi, D., Turini, F., May 2010. Data mining for
  discrimination discovery. ACM Trans. Knowl. Discov. Data 4~(2), 9:1--9:40.

\bibitem[{Sriperumbudur et~al.(2011)Sriperumbudur, Fukumizu, and
  Lanckriet}]{sriperumbudur2011universality}
Sriperumbudur, B.~K., Fukumizu, K., Lanckriet, G.~R., 2011. Universality,
  characteristic kernels and rkhs embedding of measures. Journal of Machine
  Learning Research 12~(Jul), 2389--2410.

\bibitem[{Zafar et~al.(2017)Zafar, Valera, Gomez~Rodriguez, and
  Gummadi}]{zafar2017fairness}
Zafar, M.~B., Valera, I., Gomez~Rodriguez, M., Gummadi, K.~P., 2017. Fairness
  beyond disparate treatment \& disparate impact: Learning classification
  without disparate mistreatment. In: Proceedings of the 26th International
  Conference on World Wide Web. International World Wide Web Conferences
  Steering Committee, pp. 1171--1180.

\bibitem[{Zemel et~al.(2013)Zemel, Wu, Swersky, Pitassi, and
  Dwork}]{ZemelICML13}
Zemel, R.~S., Wu, Y., Swersky, K., Pitassi, T., Dwork, C., 2013. Learning fair
  representations. In: ICML (3). Vol.~28. pp. 325--333.

\bibitem[{Zeng et~al.(2016)Zeng, Ustun, and Rudin}]{zeng2015interpretable}
Zeng, J., Ustun, B., Rudin, C., 2016. Interpretable classification models for
  recidivism prediction. Jour. of the Royal Stat. Soc.: Series A (Statistics in
  Society).

\end{thebibliography}
%}

\appendix
\section{Proof of Proposition \ref{gp_conversion}}\label{gp_conversion_proof}
Through feature maps $\phi(\cdot)$ and $\psi(\cdot)$, we map $\bfx_i$ and $\bfs_i$ into the RKHS $\mathcal{H}_k$ and $\mathcal{H}_l$ with kernel $k(\bfx,\bfy) = \phi(\bfx)^\top\phi(\bfy)$ and $l(\bfx,\bfy) = \psi(\bfx)^\top\psi(\bfy)$ respectively.\footnote{ For ease of presentation, we have abused the notation of inner product, since both  $\phi(\cdot)$ and $\psi(\cdot)$ can be infinite dimensional.} 
We form the estimation as $f(\bfx_i) = \phi(\bfx_i)^\top\bbeta$. Denote $\Phi = [\phi(\bfx_1),\cdots,\phi(\bfx_n)]^\top$ and $\Psi = [\psi(\bfs_1),\cdots,\psi(\bfs_n)]^\top$, we have that $\bff = \Phi \bbeta$. In addition, we have $\bf{K} = \Phi \Phi^\top$ and $\bf{L} = \Psi \Psi^\top$. We have shown that solving problem \eqref{fairbayesianlearning_hsic} is equivalent to solve Eq.\eqref{regprimal}, which states that the fair learning can be cast as the following optimization problem
\begin{equation}
    \min_{\beta} \bigg\{\frac{1}{\lambda}V(\bfy,\bPhi\bbeta) 
    + \bbeta^\top \bbeta 
    + \delta \bbeta^\top \bPhi^\top \HH\LL\HH \bPhi \bbeta\bigg\}.\nonumber
\end{equation}
Using the negative conditional log-likelihood as the loss, i.e. $V(f(\bfx_i),y_i)=-\log p(y_i|\bfx_i)$ and by rescaling, it can be rewritten as 
\begin{equation}\label{fair_primal}
    \min_{\beta} \bigg\{-\sum_{i=1}^n \frac{\log p(y_i|\phi(\bfx_i)^\top\bbeta)}{\lambda}  + \bbeta^\top \bbeta 
    + \delta \bbeta^\top \bPhi^\top \HH\LL\HH \bPhi \bbeta\bigg\}.
\end{equation}

Let us consider a Gaussian Process model with prior kernel defined as in the statement of Proposition \ref{gp_conversion} and the likelihood function $p$. For a set of observations $\{\bfx_i,y_i\}_{i=1}^{n}$, the model reads 
\begin{eqnarray}
\bff \sim \mathcal{N}(\bf{0}, \bf{K}-\bf{K}(\bf{K}\bf{H}\bf{L} \bf{H} + (\delta\II)^{-1})^{-1}(\HH\LL\HH)\bf{K}),\nonumber\\
p(Y|\bff) \sim \prod_{i=1}^n p(y_i|f(\bfx_i).
\end{eqnarray}
Notice that
\begin{eqnarray}
&&\bf{K}-\bf{K}(\bf{K}\bf{H}\bf{L} \bf{H} + (\delta\II)^{-1})^{-1}(\HH\LL\HH)\bf{K} \nonumber\\
&=& \bf{K}-\bf{K}(\bf{K} + \delta^{-1}(\bf{H}\bf{L} \bf{H})^{-1})^{-1}(\bf{K}+\delta^{-1}(\bf{H}\bf{L} \bf{H})^{-1})\nonumber\\
&&+\bf{K}(\bf{K} + \delta^{-1}(\bf{H}\bf{L} \bf{H})^{-1})^{-1}\delta^{-1}(\bf{H}\bf{L} \bf{H})^{-1}\nonumber\\
&=& \bf{K}(I + \delta\bf{K}\bf{H}\bf{L}\bf{H})^{-1} = (\bf{K}^{-1} + \delta \bf{H}\bf{L}\bf{H})^{-1}\label{cov_dual}
\end{eqnarray}
Hence, 
\begin{eqnarray}
\log p(\bff|Y) &\propto & \log p(Y|\bff) \cdot \log p(\bff)\nonumber\\
&\propto & \frac{1}{\lambda}\sum_{i=1}^n\log p(y_i|f(\bfx_i))\nonumber\\
&& +\log \mathcal{N}(\bf{0},(\bf{K}^{-1} + \delta \bf{H}\bf{L}\bf{H})^{-1}) \nonumber \\
&\propto &\frac{1}{\lambda}  \sum_{i=1}^n\log p(y_i|f(\bfx_i)) -\bff^\top(\bf{K}^{-1} + \delta \bf{H}\bf{L}\bf{H})\bff \nonumber\\
&=& \frac{1}{\lambda}\sum_{i=1}^n\log p(y_i|f(\bfx_i)\nonumber \\
&& -\bff^\top\bf{K}^{-1}\bff -\delta\bff^\top \bf{H}\bf{L}\bf{H}\bff\label{cov_primal}
\end{eqnarray}
Since $\bff = \bPhi \bbeta$ and $\bf{K} = \Phi\Phi^\top$, we have
\begin{eqnarray}
\eqref{cov_primal} = \frac{1}{\lambda}\sum_{i=1}^n\log p(y_i|f(\bfx_i))-\bbeta^\top\bbeta -\delta\bbeta^\top\Phi^\top\bf{H}\bf{L}\bf{H}\Phi\bbeta \label{map_prob}
\end{eqnarray}
Hence, the MAP estimate of the above probabilistic model is equivalent to the regularized ERM Problem \eqref{fair_primal}. This concludes our proof.

\end{document}